\documentclass{article}

\usepackage{xcolor}
\usepackage[preprint]{corl_2026} % Uncomment for pre-prints (e.g., arxiv); This is like ``final'', but will remove the CORL footnote.
\definecolor{RokenAccent}{HTML}{8A1F2D}
\definecolor{RokenAccentDark}{HTML}{641722}
\definecolor{RokenSoftBg}{HTML}{FBF1F3}
\definecolor{RokenDivider}{HTML}{E6D7DA}

\hypersetup{
  pdftitle={Robots as Tokens: Unified Diffusion Transformer for Coordinated Multi-Robot Trajectory Generation},
  pdfauthor={Ruofei Bai, Jie Chen, Yuxin Cai, Jun Li, Wei-Yun Yau, Lihua Xie},
  pdfkeywords={Multi-Robot Planning, Diffusion Models, Trajectory Generation},
  linkcolor=RokenAccentDark,
  citecolor=RokenAccent,
  filecolor=RokenAccentDark,
  urlcolor=RokenAccent
}

\usepackage[T1]{fontenc}
\usepackage{amsmath}
\usepackage{amssymb}
\usepackage{graphicx}
\usepackage{wrapfig}
\usepackage{capt-of}
\usepackage{subcaption}
\usepackage{multirow}
\usepackage{enumitem}
\usepackage{placeins}


\usepackage{algorithm}
\usepackage{algpseudocode}

\newcommand{\affmark}[1]{\textsuperscript{\textcolor{RokenAccentDark}{#1}}}

\title{\texorpdfstring{\textcolor{RokenAccentDark}{Robots as Tokens}: Unified Diffusion Transformer for Coordinated Multi-Robot Trajectory Generation}{Robots as Tokens: Unified Diffusion Transformer for Coordinated Multi-Robot Trajectory Generation}}

\author{
  Ruofei Bai\affmark{1,2}, Jie Chen\affmark{3}, Yuxin Cai\affmark{1,2}, Jun Li\affmark{2}, Wei-Yun Yau\affmark{2}, Lihua Xie\affmark{1}\\
  \affmark{1} Nanyang Technological University, Singapore \\
  \affmark{2} Agency for Science, Technology and Research, Singapore \\
  \affmark{3} National University of Singapore, Singapore
}

\begin{document}
\maketitle

%===============================================================================

\begin{abstract}
The success of generative models in language, image, and video generation has inspired extensive applications to generative robot planning. 
However, most existing works either focus on single-robot planning, or generate multi-robot trajectories in a sequential manner with iterative post-processing to resolve inter-robot conflicts.
% Generative models have been predominant in language, image, and video generation. 
In this work, we investigate whether coordinated multi-robot trajectories, as a special spatiotemporal distribution, can be learned and generated with a generative model in a feed-forward manner. 
% Previous generative planning methods either rely on sequential individual planning, or on iterative post-processing to resolve inter-robot conflicts.
We propose \textbf{Ro}bots as To\textbf{ken}s (Roken), a unified diffusion transformer that directly generates multi-robot trajectories that satisfy both (individual) safety and (global) connectivity constraints. 
% The core design of Roken is representing each robot as a token, that natually interacts with each other through self-attention, and cross-attend to map tokens encoded from a common occupancy map for individual environment awareness.
% However, we found that a simple denoising loss is insufficient for learning the conditional multi-robot trajectories.
% We ground the reason in Bayes formula and introduce several auxiliary tasks to provide multi-scale spatial-temporal self-supervision for learning the conditional distribution, including local occupancy reconstruction and long-horizon waypoint prediction. 
The core design of Roken is to represent each robot as a discrete token, allowing them to naturally interact with each other through self-attention, and cross-attend to map tokens for environment layouts. 
We further introduce several auxiliary tasks based on Bayes' theorem, specifically local occupancy reconstruction and long-horizon waypoint prediction, to provide multi-scale spatial-temporal supervision for efficient learning of the conditional distribution.
In training, Roken absorbs diverse expert trajectories from different team sizes.
During inference, Roken behaves as a versatile multi-robot planner that can handle single-robot planning, coordinated multi-robot trajectory generation, and conditional trajectory generation by fixing some robot tokens as conditions.
% We design Roken following a pure data-driven paradigm, without explicit connectivity graph contruction or communication constraints modeling as in previous work, making the framework more transferable across different scenarios.
Experiments in diverse cluttered environments show that Roken can generate coordinated multi-robot trajectories to perform connectivity-constrained goal navigation tasks with high success rates, outperforming the baseline method used to generate the training dataset.
Roken also demonstrates good scalability after training with mixed team sizes, and shows generalization to unseen or partially observed environments, verifying its potential as a unified multi-robot planner that can learn from diverse data and perform versatile tasks.
% Generative models have shown strong capability in modeling complex distributions in language, images, video, and control. Inspired by this capability, we study coordinated multi-robot trajectory planning as a conditional generation problem: given an occupancy map, robot starts, and goal conditions, the model should generate a set of trajectories that reaches the specified goals while avoiding collisions and maintaining team communication. 
\end{abstract}


\begin{figure}[h]
    \centering
    \includegraphics[width=\linewidth]{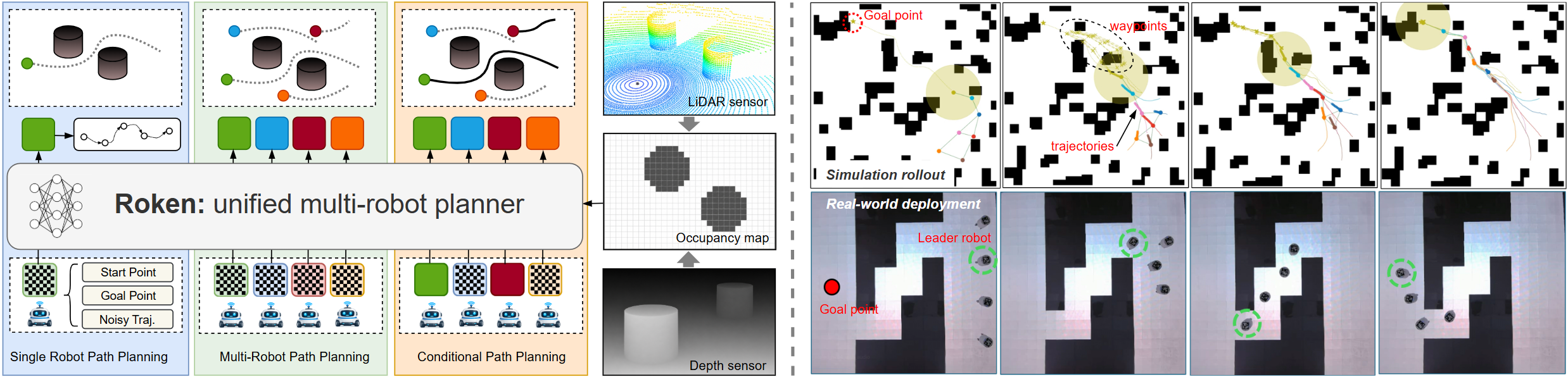}
    \vspace{-5pt}
    \caption{(Left) Roken supports three planning modes with the same robot-token diffusion model: single-robot planning, coordinated multi-robot planning, and conditional planning with fixed robot tokens. (Right) Deployment of Roken in a cluttered simulation environment with seven robots, and in a real-world maze environment with four robots, all with a single Roken model.}
    \vspace{-20pt}
    \label{fig:intro_capabilities}
\end{figure}

%===============================================================================

\section{Introduction}

% Multi-robot systems have broad applications in search and rescue, warehouse automation, and environmental monitoring.
Multi-robot systems have broad applications in search and rescue~\cite{tian2020search, wang2024viper}, autonomous exploration~\cite{bai2024graph, chiun2025marvel}, and environmental monitoring~\cite{xu2024cost}.
To ensure effective coordination, it is essential for the robots to maintain an underlying connected communication network for real-time information sharing and decision-making.
However, such connectivity constraints usually conflict with robots' navigation tasks, as the communication links can be easily broken when robots navigate in obstacle-cluttered environments.
In such a case, it is challenging to plan coordinated trajectories that not only achieve navigation tasks but also satisfy both safety and connectivity maintenance constraints.

Existing methods consider either continuous-time or discrete-time connectivity maintenance. 
The former are usually based on potential fields or optimization-based methods~\cite{robuffo2013passivity, nestmeyer_DecentralizedSimultaneous_2017}, while the latter rely on discrete candidate sampling and combinatorial evaluation~\cite{bai2026lineofsightconstrainedmultirobotmaplessnavigation,  shi_CommunicationAwareMultirobot_2021}.
% which require continuous connectivity maintenance and can easily get stuck in local minima areas due to confliction of multiple components, including collision aovidance, goal navigation, and communication maintenance.
% Exising methods tackle this problem in either continuous or discrete settings. 
% The continuous appraochs require continuous connectivity maintenance, and is usually based on potential fields, or optimization-based methods. 
% The discrete formulation is usually based on graph search, where candidate future positions of each robot are first sampled, and then combined with other robots candiates for combinatorial evaluation. 
However, in cluttered environments, the gradient field-based methods can easily get stuck in local minima areas due to confliction of multiple components, including collision avoidance, goal navigation, and communication maintenance. 
The discrete approaches suffer from the curse of dimensionality for candidates evlauation, and often require downstream repair to generate collision-free trajectories~\cite{shi_CommunicationAwareMultirobot_2021, ishat2024fast}.

Recently, generative models have shown strong capability in modeling complex distributions like language, images and videos~\cite{ho2020denoisingdiffusionprobabilisticmodels, rombach2022highresolutionimagesynthesislatent, blattmann2023stablevideodiffusionscaling, wan2025wanopenadvancedlargescale}, and they have been applied to generative trajectory planning in robot navigation and manipulation tasks~\cite{janner_PlanningDiffusionFlexible_2022, chi2024diffusionpolicyvisuomotorpolicy, sridhar2023nomadgoalmaskeddiffusion, black2026pi0visionlanguageactionflowmodel, chen2026imaginav}.
However, most diffusion planning works have focused on single robot planning, or fixed-environment trajectory generation~\cite{janner_PlanningDiffusionFlexible_2022}.
The individually generated trajectory either needs to be conditioned on existing robot plans, or requires further iterative refinement to satisfy multi-robot constraints~\cite{shaoul_MultiRobotMotion_2024,liang_SimultaneousMultiRobot_2025,liang_DiscreteGuidedDiffusion_2025}.
% Therefore, it is non-trivial to extend these methods to multi-robot planning and broader environments.
This motivates us to ask the question: can we learn a generative model that directly generates coordinated multi-robot trajectories in a feed-forward manner, and scales to varying numbers of robots?

In this work, we propose Robots as Tokens (Roken), a unified generative model based on diffusion transformer (DiT)~\cite{peebles_ScalableDiffusionModels_2023, vaswani2017attention} that treats coordinated multi-robot trajectories as a special spatiotemporal distribution, and learns the map-conditioned trajectory distribution from expert data.
The key design of Roken is to represent each robot as a token input to the DiT model, where the transformer structure naturally allows robot tokens to interact with each other through self-attention, and attend to environment map tokens for spatial awareness.
In training, Roken absorbs expert trajectories with different team sizes as different numbers of robot tokens, and supports conditional training by treating some robot tokens as known conditions.
During inference, Roken behaves as a versatile multi-robot planner that can handle single-robot planning, coordinated multi-robot planning, and conditional planning by simply changing the number of robot tokens and their trajectory masks, without any change to the model architecture or parameters.
This mimics the versatility of human planners, who can not only play Ludo with only one role, but also play chess with many different roles and opponents' conditions.
The contributions of this paper are:
\begin{itemize}[leftmargin=*, itemsep=0.2em]
    \item We propose Roken, a unified diffusion transformer that encodes each robot as a token, learns and generates communication-constrained coordinated trajectories of varying numbers of robots in a feed-forward manner, without sequential individual planning or iterative post-processing.
    \item We design multi-scale spatial-temporal self-supervision through auxiliary tasks for conditional distribution learning, enhancing the condition injection for the model's map awareness, long-range planning, and collision avoidance behaviors.
    \item We conduct extensive experiments to verify the capabilities, scalability, and generalizability of Roken in diverse complex maps, and demonstrate its deployment in real-world navigation tasks.
\end{itemize}

% Each robot is represented as a token containing a noisy future trajectory, its start position, a relative goal condition, a goal mask, and a trajectory mask. Robot tokens interact through self-attention, while map patch tokens are queried through cross-attention. This architecture treats the team as an unordered but interacting set and keeps the number of robot tokens explicit, which makes it natural to switch between single-robot planning, full-team planning, and conditional generation.

% An important design choice is that Roken does not impose communication maintenance through an explicit differentiable connectivity objective during imitation learning. For example, one could build gradients from the Fiedler eigenvalue $\lambda_2$ of the communication-graph Laplacian and directly optimize robot positions to increase algebraic connectivity. Instead, we treat connectivity maintenance as a behavioral structure contained in expert coordinated trajectories. The model learns this structure in a purely data-driven manner through robot-token interactions, without back-propagating handcrafted connectivity gradients through the imitation objective.

\section{Related Work}

\subsection{Multi-Robot Planning and Connectivity Maintenance}

%In contrast, existing methods either rely on manual checking of connectivity constraints, or apply $\lambda_2$-based gradients to directly optimize robot positions to increase algebraic connectivity.

Multi-robot navigation under communication constraints requires balancing navigation tasks with global connectivity maintenance. Previous gradient-based methods usually encode communication and collision constraints as potential functions in a connectivity controller~\cite{robuffo2013passivity, nestmeyer_DecentralizedSimultaneous_2017, bai2025realm}. While mathematically elegant, they are inherently reactive and suffer from severe local minima in cluttered environments where navigation and connectivity objectives conflict. Alternatively, communication-aware coordination can be formulated as a combinatorial planning problem by searching for optimal future connected topologies from discrete candidates~\cite{shi_CommunicationAwareMultirobot_2021}. 
However, this discretization introduces coarse motions, often requiring complex downstream smoothing or trajectory repair to generate executable and collision-free paths.
Recent works have applied reinforcement learning (RL) to connectivity-aware navigation.
The decision space is usually simplified to discrete waypoints selection~\cite{li_DecentralizedGlobal_2022} or candidate topologies scoring~\cite{tang_DecentralizedCommunicationMaintained_2024}, leaving continuous execution to external low-level controllers.
Moreover, the coordinated behaviors are usually learned through explicit gradient guidance from constarints formulations, which makes it difficult to scale their framework to other collaborative tasks.
These limitations motivate us to design a pure data-driven learning framework that directly learns the conditional trajectory distribution from expert demonstrations, without explicit connectivity constraints modeling or gradient injection, making the framework more transferable across different scenarios.

\subsection{Diffusion Models for Trajectory Generation}

Diffusion models, known to be excellent at modeling complex distributions, have revolutionized image synthesis~\cite{ho2020denoisingdiffusionprobabilisticmodels, rombach2022highresolutionimagesynthesislatent}, video generation~\cite{blattmann2023stablevideodiffusionscaling, wan2025wanopenadvancedlargescale}, and robot planning~\cite{janner_PlanningDiffusionFlexible_2022, chi2024diffusionpolicyvisuomotorpolicy, sridhar2023nomadgoalmaskeddiffusion, black2026pi0visionlanguageactionflowmodel}. While many existing methods focus on single-agent control, diffusion models have recently been applied to generative multi-robot motion planning~\cite{shaoul_MultiRobotMotion_2024, liang_SimultaneousMultiRobot_2025,liang_DiscreteGuidedDiffusion_2025}.
Despite their significant strides, existing diffusion-based planners predominantly rely on external mechanisms to ensure trajectory feasibility. 
They typically depend on computationally expensive iterative projections to repair constraint violations~\cite{liang_SimultaneousMultiRobot_2025, liang_DiscreteGuidedDiffusion_2025}, or require coupling with discrete search solvers for combinatorial guidance~\cite{shaoul_MultiRobotMotion_2024, liang_DiscreteGuidedDiffusion_2025}. 
Different from existing methods, this work presents a unified model that learns coordinated multi-robot trajectories as a joint spatiotemporal distribution, and directly generates multi-robot trajectories in a feed-forward manner, without any iterative refinement, which is more efficient and end-to-end learnable.
We provide the global occupancy map as a common conditioning for trajectory generation instead of ego-centric observations as in single-robot settings, which requires each robot token to actively query the relevant information but improves generalizability across different sensors.

% Roken overcomes these limitations by generating safe, globally connected trajectories in a direct, feed-forward denoising manner.
% By leveraging auxiliary tasks, Roken internalizes environment awareness and learns complex constraint adherence purely from the underlying data distribution

% These methods demonstrate the promise of diffusion for multi-robot planning.
% However, they often rely on iterative constraint projection, external discrete guidance, or repair mechanisms. In contrast, Roken targets a unified feed-forward denoising model that learns coordinated multi-robot trajectory structure directly from data.
% Our work differs from the above methods in following aspects. 
% First, we defined a unified model that directly generates multi-robot trajectories. Second, 
% Third, we design the model based on purely data-driven principles, without explicit gradient injection for connectivity maintenance, making the framework more transferable across different scenarios.

\section{Preliminaries and Problem Formulation}

\subsection{Communication Connectivity}
\label{sec:communication_graph}
Let $\mathcal{N}=\{1,\ldots,N\}$ be a team of robots that navigate in an environment $\mathcal{W}$ with obstacles. 
Two robots are considered to be in communication if they are within a certain distance $R$ of each other, i.e., $\|p^i_t - p^j_t\|_2 \leq R$, where $p^i_t$ denotes the position of robot $i$ at time $t$.
The communication graph at time $t$ is defined as $\mathcal{G}_t = \langle\mathcal{V}_t, \mathcal{E}_t \rangle$, where $\mathcal{V}_t=\{p^i_t \mid i \in \mathcal{N}\}$ is the set of robots, and $\mathcal{E}_t = \{(i,j) \mid \|p^i_t - p^j_t\|_2 \leq R\}$ is the set of connected edges.
Given the communication graph $\mathcal{G}_t$, its graph Laplacian matrix can be obtained as $\mathcal{L}_{\mathcal{G}_t} = D - A$, where $A$ is the adjacency matrix of $\mathcal{G}_t$ and $D$ is the diagonal degree matrix.
The graph $\mathcal{G}_t$ is connected if there is a path between any two vertices in the graph, or equivalently, the second-smallest eigenvalue $\lambda_2$ of $\mathcal{L}_{\mathcal{G}_t}$ is positive, i.e., $\lambda_2(\mathcal{L}_{\mathcal{G}_t}) > 0$.
Otherwise, $\lambda_2(\mathcal{L}_{\mathcal{G}_t}) = 0$ if $\mathcal{G}_t$ is disconnected~\cite{fiedler1973algebraic}.
The gradient of $\lambda_2$ with respect to robot positions can be derived to provide directions for increasing connectivity, which is commonly used in gradient-based connectivity maintenance methods~\cite{robuffo2013passivity, nestmeyer_DecentralizedSimultaneous_2017, bai2025realm}.

\subsection{Problem Formulation}
\label{sec:constraints}

The environment is represented as a 2D occupancy map $\mathcal{H}_{\mathrm{map}} \in \{0,1\}^{H \times W}$, where occupied cells correspond to obstacles and others represent free space.
There are $N$ robots navigating in the environment, where each robot $i$ has an initial position $p^i_0 \in \mathbb{R}^2$.
One leader robot will be given a goal position $g \in \mathbb{R}^2$.
The trajectory of a robot $i$ is represented as 
$\tau^i = \{p^i_1,\ldots,p^i_T\}, \quad i \in \mathcal{N}$,
where $p^i_t$ is the position of robot $i$ at time $t$. 
The team trajectories are represented as $\mathcal{T}=\{\tau^i\}_{i=1}^N$.
The coordinated navigation task requires robots to satisfy the following constraints:
(1) \textbf{Goal-reaching}: the leader robot should reach the goal position within a specified tolerance $\epsilon_g$, i.e., $\|p^i_T - g\|_2 \leq \epsilon_g$ for the robot $i$ assigned with the goal;
(2) \textbf{Collision avoidance}: all robots should avoid collisions with obstacles and each other, i.e., $d_{\mathrm{obs}}(p^i_t) \geq d_{\mathrm{obs}}^{\min}$ and $\|p^i_t - p^j_t\|_2 \geq d_{\mathrm{coll}}^{\min}$ for all $i \neq j$ and all $t$, where $d_{\mathrm{obs}}(p)$ is the distance from position $p$ to the nearest obstacle, and $d_{\mathrm{obs}}^{\min}$ and $d_{\mathrm{coll}}^{\min}$ are safety thresholds;
(3) \textbf{Communication maintenance}: the team should maintain communication connectivity, i.e., $\lambda_2(\mathcal{L}_{\mathcal{G}_t}) > 0$ for all $t$.

% \begin{itemize}
%     \item \textbf{Goal-reaching}: the leader robot should reach the goal position within a specified tolerance $\epsilon_g$, i.e., $\|p^i_T - g\|_2 \leq \epsilon_g$ for the robot $i$ assigned with the goal;
%     \item \textbf{Collision avoidance}: all robots should avoid collisions with obstacles and each other, i.e., $d_{\mathrm{obs}}(p^i_t) \geq d_{\mathrm{obs}}^{\min}$ and $\|p^i_t - p^j_t\|_2 \geq d_{\mathrm{coll}}^{\min}$ for all $i \neq j$ and all $t$, where $d_{\mathrm{obs}}(p)$ is the distance from position $p$ to the nearest obstacle, and $d_{\mathrm{obs}}^{\min}$ and $d_{\mathrm{coll}}^{\min}$ are safety thresholds;
%     \item \textbf{Communication maintenance}: the team should maintain communication connectivity, meaning that the communication graph $\mathcal{G}_t$ formed by the robots' positions at time $t$ should remain connected for all $t$, i.e., $\lambda_2(\mathcal{L}_{\mathcal{G}_t}) > 0$ for all $t$.
% \end{itemize}

\subsection{Dataset Collection}

We collect the training dataset $\mathcal{D}$ with diverse team sizes using a classical potential field-based method \textbf{Laplacian} adopted from~\cite{nestmeyer_DecentralizedSimultaneous_2017} based on the graph Laplacian.
The dataset distribution is shown in Fig.~\ref{fig:dataset}.
To provide diverse training environments, we use a 2D occupancy map dataset from the 2d-path-planning-dataset\footnote{https://www.kaggle.com/datasets/dcaffo/2dpathplanningdataset}, and select 5000 maps from the dataset for training and 1000 maps for evaluation. 
During dataset collection, the \textbf{Laplacian} method first randomly samples start and goal positions, and lets the leader robot follow an optimal A* path while others maintain connectivity. 
The team only maintains necessary connections with topology optimization following~\cite{bai2026lineofsightconstrainedmultirobotmaplessnavigation}.
We smooth the expert trajectories to mitigate the discontinuity, and augment the dataset with random rotation to increase its diversity.
During training, we randomly sample a fixed-length multi-robot action chunk and a goal position along the future trajectory of the leader robot from an episode.
Note that we abuse the notation $\mathcal{T}$ to represent either the full trajectory or the relative action chunk for notation convenience.

% Each data episode is defined as $\langle \mathcal{H}_{\mathrm{map}}, \mathcal{P}_{0}, g, \mathcal{T} \rangle$, where  $\mathcal{P}_0 = \{p^i_0\}_{i=1}^N$ is the set of initial positions for all robots.
% The trajectories $\mathcal{T}$ are smoothed to mitigate the discontinuity caused by the baseline method.

% Moreover, we augmented the dataset with random rotation to increase the diversity of the dataset.

\section{Methodology}

% Different from existing works, this work aims to establish a pure data-driven paradigm for learning multi-robot coordinated behaviors, for potential scaling and generalization. 
% We do not explicitly impose connectivity constraints during both imitation learning and reinforcement learning stages, but instead learning from expert demonstrations and reward signals.

\subsection{Model Structure}
\label{sec:model_structure}

\begin{figure}[t]
    \centering
    \begin{subfigure}[t]{.71\linewidth}
        \centering
        \includegraphics[width=\linewidth]{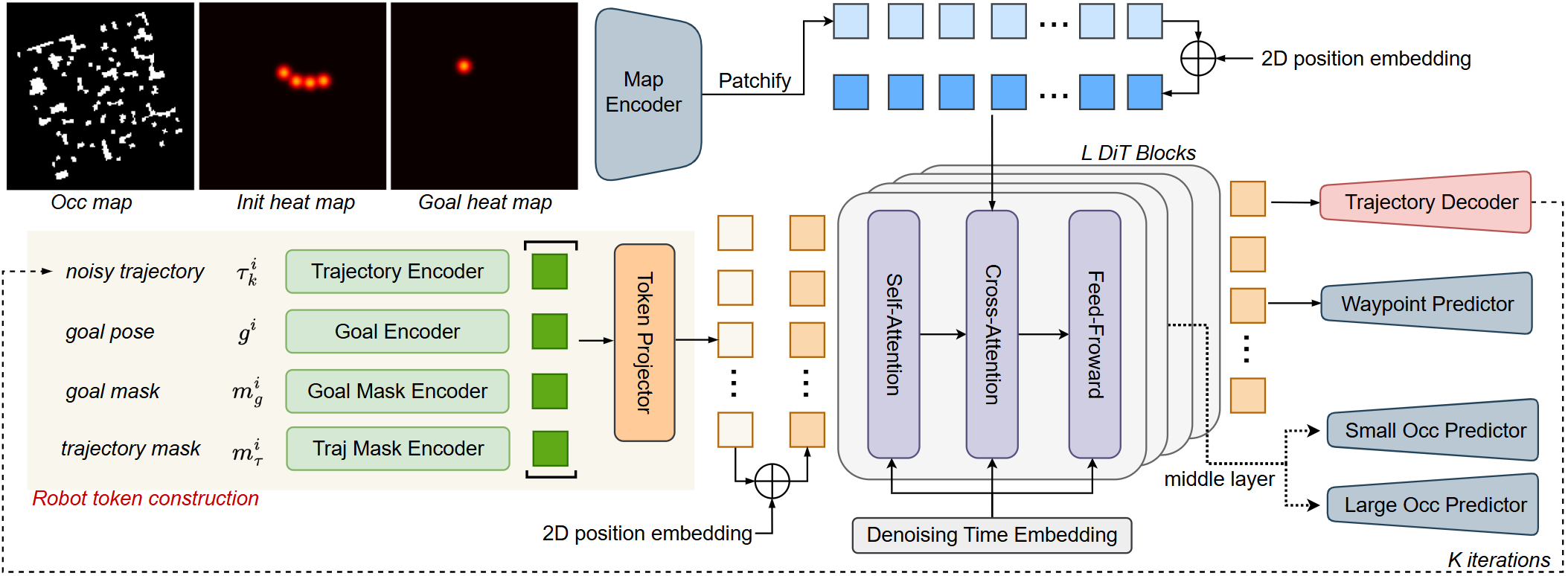}
        \caption{Network Architecture of Roken.}
        \label{fig:framework_architecture}
    \end{subfigure}\hfill
    \begin{subfigure}[t]{.26\linewidth}
        \centering
        \includegraphics[width=\linewidth]{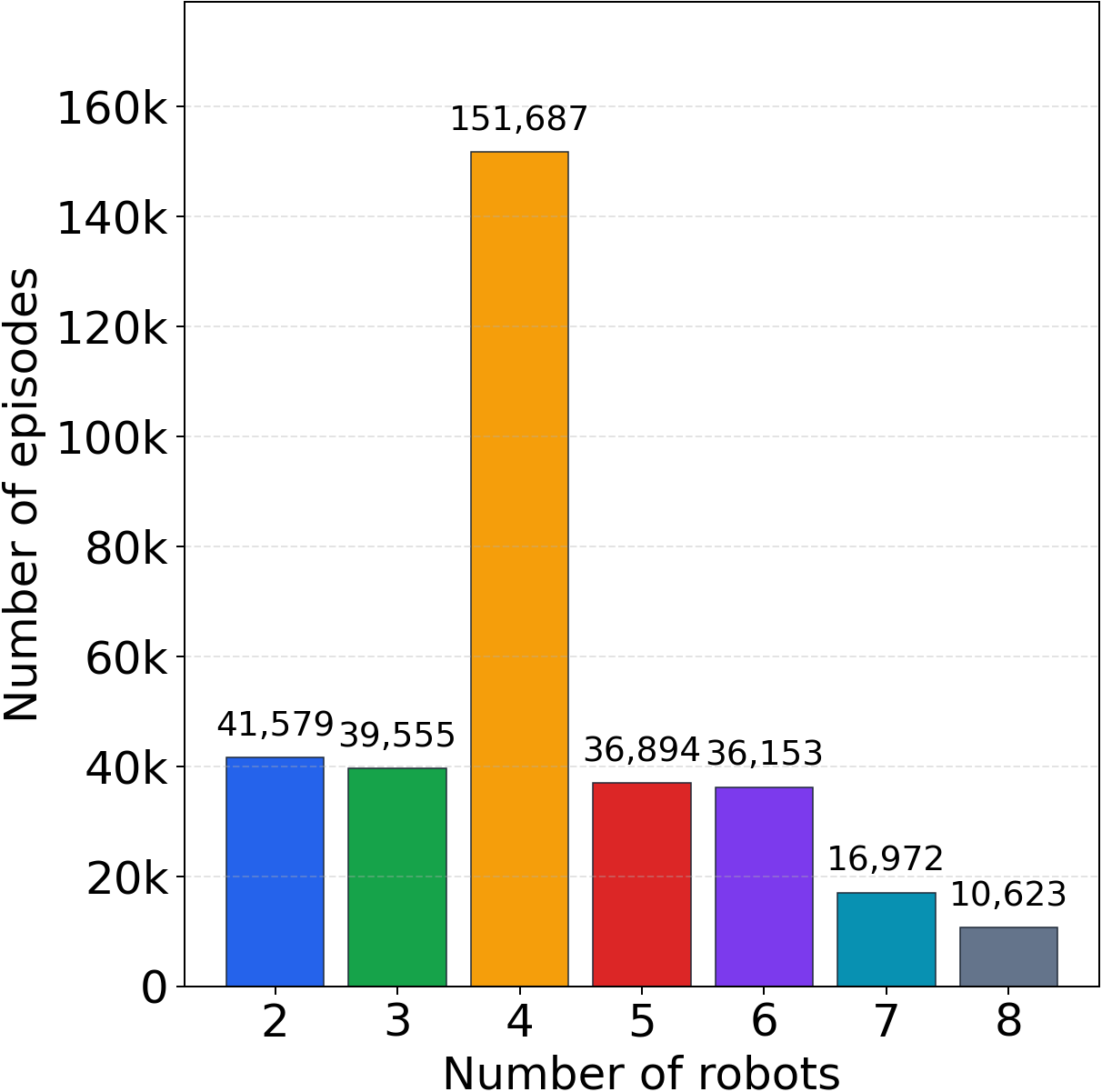}
        \caption{Dataset distribution.}
        \label{fig:dataset}
    \end{subfigure}
    \vspace{-5pt}
    \caption{Overview of the proposed Roken framework and the training dataset distribution across team sizes.}
    \label{fig:framework}
    \vspace{-15pt}
\end{figure}

We first give an overview of the proposed neural network architecture, called Roken, as shown in Fig.~\ref{fig:framework}(a). 
Roken is an end-to-end diffusion transformer, which takes multiple \emph{robot tokens} as inputs, and passes them through  multiple transformer decoder layers while cross-attending to \emph{map tokens} to extract occupancy information, and outputs latent vectors which are then decoded into clean trajectories.
The overall training objective is to maximize the conditional likelihood of feasible coordinated trajectories:
\begin{equation}
    \max_{\theta}
    \mathbb{E}_{\mathcal{T}, \mathcal{H}_{\mathrm{map}}, \mathcal{P}_0, g \sim p_{\mathrm{data}}}
    \left[\log p_\theta(\mathcal{T}\vert \mathcal{H}_{\mathrm{map}}, \mathcal{P}_0, g)\right].
\label{eq:objective}
\end{equation}
% where $\mathcal{T}$ is the set of trajectories for all robots, $\mathcal{H}_{\mathrm{map}}$ is the occupancy map, $\mathcal{P}_0 = \{p^i_0\}_{i=1}^N$ is the set of initial positions for all robots, and $g\in \mathbb{R}^2$ is the goal position for the leader robot.

\textbf{Robot Tokens:} 
We introduce the novel concept of robot tokens in this work for scalable multi-robot trajectory generation. 
Each robot token is a latent vector that encapsulates the information of a single robot, including its noisy future action chunk and goal information, defined as
% The robot tokens interact with each other through self-attention, and cross-attends to map tokens for occupancy information.
% three advantages: 1. accepts varying number of robots; 2. permutation-invariant to robot order; 3. allow flexible training that absorbing expert trajectories with different team sizes, and conditional training. 
% Different from existing work that generates action chunks with each token representing the action at a single time step, our design makes changing the team size as simple as changing the number of robot tokens inputed to the model.
% The robot tokens then attend to each other through self-attention, and attend to map patch tokens through cross-attention, and finally are used to decode the predicted clean trajectories.
% Specifically, the robot token for a robot $i$ is constructed by embedding its noisy action chunk, relative goal, goal mask, and trajectory mask:
\begin{equation}
    z^i_0 = \phi_{\mathrm{tok}}\left(
    \operatorname{CAT}\left[
    \phi_{\tau}(\tau^i_k),
    \phi_g(g^i-p^i_0),
    \phi_{mg}(m^i_g),
    \phi_{m\tau}(m^i_\tau)
    \right]\right) + \operatorname{PE}(p^i_0),
\end{equation}
where $\phi_{\tau}$, $\phi_g$, $\phi_{mg}$, and $\phi_{m\tau}$ are embedding layers for the noisy action chunk $\tau^i_k$ at the $k$-th denoisingtime step, relative goal $g^i-p^i_0$, goal mask $m^i_g\in \{0,1\}$, and trajectory mask $m^i_\tau\in \{0,1\}$, respectively.
The goal mask indicates the leader robot; and the trajectory mask indicates whether the action chunk is generated from noise or kept fixed as a condition for conditional generation.
The linear projection $\phi_{\mathrm{tok}}$ maps the concatenated embeddings into the robot token space, and adds sinusoidal positional embeddings $\operatorname{PE}(p^i_0)$.
All input robot tokens are denoted as $Z_0 = \{z^i_0\}_{i=1}^N$.

With robot tokens, our unified DiT model Roken enjoys three advantages: (1) it can accept varying numbers of robots by simply changing the number of robot tokens; (2) it is permutation-invariant to the order of robot tokens, which is natural for multi-robot systems; and (3) it allows flexible training with diverse datasets containing different team sizes, as well as conditional training by treating some robot tokens as known conditions and others to be generated from noise.

% This tokenization enables Roken to handle different team sizes by changing the number of robot tokens and using padding masks up to a maximum robot count, in both training and inference. 
% It also supports conditional generation by treating some robot tokens as known conditions (with trajectory masks indicating no noise) and others as unknowns to be generated from noise.

\textbf{Map tokens:} 
% As we provide a common occupancy map for all robots, each robot needs to localize itself w.r.t. the map, and actively query the map for relevant obstacle information.
% In our model, the map is represented as spatial patch tokens that are queried by robot tokens through cross-attention.
To help the ResNet-based map encoder learn relevant representations for robots rather than a general map compression, we extend the map tensor with two additional channels: a robot-start heatmap and a goal heatmap.
A Gaussian kernel is placed at each robot's start position and at the goal position in their respective channels, providing spatial anchors for the map encoder to better compress the map information.
After the map encoder, the downsampled map is divided into patch tokens with positional embeddings from the patch center coordinates.
% The multichannel map tensor first passed through a  map encoder, and then is projected into patch tokens through a patch embedding layer.
% The normalized center coordinate of each patch are encoded through 2-D positional embeddings, and added to each map token as the final input to the transformer decoder.
Robot tokens attend to these map tokens through transformer decoder cross-attention, with
$    Z_{\ell+1} =
    \mathrm{DecoderLayer}_{\ell}(Z_\ell, M, e_k)$,
where $Z_\ell = \{z^i_\ell\}_{i=1}^N$ are $N$ input latent embeddings at the $\ell$-th layer, $M = \{h^1, \ldots, h^m\}$ are $m$ map patch tokens, and $e_k$ is the diffusion timestep embedding. 
The denoising time step embedding is injected into the network with AdaLN-Zero~\cite{peebles_ScalableDiffusionModels_2023} for stable diffusion transformer training.

% \textbf{Time Conditioning:} The final model variant uses AdaLN-Zero-style timestep modulation for stable diffusion transformer training~\cite{peebles_ScalableDiffusionModels_2023}. 
% The the default zero-gate design in AdaLN-Zero is disabled for the cross-attention output, which allows the map information to enter the robot-token stream from the early stages of training.
% We found the default zero-gate design in AdaLN-Zero makes it difficult for the robot tokens to learn to query the map information through cross-attention.
% In our implementation, we remove the zero gate for the cross-attention output, allowing map information to enter the robot-token stream from the early stages of training.

\textbf{Auxiliary Heads:} 
% Roken includes three types of output heads: a main trajectory decoder head that decodes the latent output into predicted noises; and two auxiliary heads for local occupancy reconstruction and waypoint prediction.
While it is straightforward to use diffusion-based imitation learning to train the trajectory decoder head, we found that this alone is difficult for effective learning of the map-conditioned trajectory distribution.
The main reason is that the relative goal position $g - p^i_0$ in robot token $z^i_0$ provides a strong shortcut for the model to simply learn to move towards the goal, without querying the map information.
% We interpret the second reason following the Bayes-rule.
To show that, we can rewrite the learning objective in Eq.~\eqref{eq:objective} following the Bayesian formula as follows:
% Specifically, the learning objective can be viewed as maximizing the log-likelihood of trajectories given the map information\footnote{Note that we cannot treat the map and trajectories as a combined distribution $p(\mathcal{T}, \mathcal{H}_{\mathrm{map}})$, because for each single map, the expert trajectories in the dataset only cover a small subset of the feasible trajectory space, and the model can easily overfit to this small subset without the generalizability to other task configurations in the same map.} $\log p(\mathcal{T}\mid \mathcal{H}_{\mathrm{map}})$.
% Following the Baysian formula, we have:
% \begin{equation}
% \label{eq:bayes_log_cond_traj}
% \log p(\mathcal{T}\mid \mathcal{H}_{\mathrm{map}})
% = \log p(\mathcal{T}) + \log p(\mathcal{H}_{\mathrm{map}}\mid \mathcal{T}) - \log p(\mathcal{H}_{\mathrm{map}}).
% \end{equation}
\begin{equation}
% \small
\label{eq:bayes_log_cond_traj}
\log p(\mathcal{T}\mid \mathcal{H}_{\mathrm{map}}, \mathcal{P}_0, g)
= \log p(\mathcal{T}\mid \mathcal{P}_0, g)
+ \log p(\mathcal{H}_{\mathrm{map}} \mid \mathcal{T}, \mathcal{P}_0, g)
- \log p(\mathcal{H}_{\mathrm{map}}\mid \mathcal{P}_0, g).
\end{equation}
Due to the existence of the training shortcut, the model actually learns $p(\mathcal{T} \mid \mathcal{P}_0, g)$, rather than the desired conditional distribution $p(\mathcal{T} \mid \mathcal{H}_{\mathrm{map}}, \mathcal{P}_0, g)$.
% In Eq.~\eqref{eq:bayes_log_cond_traj}, the term $\log p(\mathcal{H}_{\mathrm{map}}\mid \mathcal{P}_0, g)$ can be treated as a constant. 
% The denoising loss directly supervises the learning of $\log p(\mathcal{T} \mid \mathcal{P}_0, g)$, but omits $\log p(\mathcal{H}_{\mathrm{map}}\mid \mathcal{T}, \mathcal{P}_0, g)$ that measures whether a trajectory is compatible with the surrounding environment.
% \footnote{Note that we cannot treat the map and trajectories as a combined distribution $p(\mathcal{T}, \mathcal{H}_{\mathrm{map}}, \mathcal{P}_0, g)$, because for each single map, the expert trajectories in the dataset only cover a small subset of the feasible trajectory space, and the model can easily overfit to this small subset without the generalizability to other task configurations in the same map.}
Therefore, we introduce the \emph{local occupancy reconstruction} heads to enhance the learning of the second term $p(\mathcal{H}_{\mathrm{map}} \mid \mathcal{T}, \mathcal{P}_0, g)$, thereby supplementing the learning of the target conditional distribution in Eq.~\eqref{eq:bayes_log_cond_traj}. 
Note that the term $\log p(\mathcal{H}_{\mathrm{map}}\mid \mathcal{P}_0, g)$ is unrelated to $\mathcal{T}$ and can be treated as a constant. 
The occupancy reconstruction head asks each middle-layer latent vector to decode two local occupancy maps with both original and large patch scales, centered around the corresponding robot position.
This encourages the robot token to localize itself within the map, aggregate surrounding discrete map patches, and understand both fine-grained and large-scale obstacle layouts for trajectory planning.

% the occupancy map (with both small and large patch scales) centered around the corresponding robot position, which encourages the robot token to localize itself within the map, aggregate surrounding discrete map patchs, and understand the obstacle layout for trajectory planning.

% Specifically, each middle-layer latent token of Roken is output to an auxiliary head to decode the occupancy map (with both small and large patch scales) centered around the corresponding robot position.
% We use the middle-layer latent vectors to avoid capacity competition with the main trajectory generation task.
% \footnote{We use the middle-layer latent vectors to avoid capacity competition with the main trajectory generation task, and only use it to guide the representation learning of the conditional distribution.}
% It encourages the model to learn to query the discrete map patch tokens, aggregate them, and understand the local obstacle layout for collision avoidance.
% We found that this auxiliary task significantly improves the obstacle aovidance behavior of the geneated trajectories, verying our analysis on conditional distribution learning.

We also introduce another auxiliary \emph{waypoint prediction} head to enhance the long-range planning capability of the model, which predicts a few sparse but farther waypoints along the trajectory.
Together, the two auxiliary heads form a multi-scale spatial-temporal supervision for the model.
They are only used during training and are disabled during inference.

\subsection{Training Objectives}
\label{sec:training_objectives}

The training objective in Eq.~\eqref{eq:objective} is optimized through a denoising score-matching loss with several auxiliary losses.
The overall loss function is defined as:
\begin{equation}
\mathcal{L}
=
\lambda_{1}\mathcal{L}_{\mathrm{traj}}
+ \lambda_{2}\mathcal{L}_{\mathrm{wp}}
+ \lambda_{3}\mathcal{L}_{\mathrm{occ}} 
+ \lambda_{4}\mathcal{L}_{\mathrm{sdf}},
\label{eq:overall_loss}
\end{equation}
where 
$    \mathcal{L}_{\mathrm{traj}} =
    \mathbb{E}\left[
    \left\|
    \epsilon -
    \epsilon_{\theta}(\mathcal{T}_k, \mathcal{P}_0, g, \mathcal{H}_{\mathrm{map}}, k)
    \right\|_2^2
    \right]$ is the trajectory denoising loss, 
% \begin{equation}
%     \mathcal{L}_{\mathrm{traj}} =
%     \mathbb{E}\left[
%     w_{\tau}
%     \left\|
%     \epsilon -
%     \epsilon_{\theta}(\mathcal{T}_k, \mathcal{P}_0, g, \mathcal{H}_{\mathrm{map}}, k)
%     \right\|_2^2
%     \right],
% \end{equation}
$\epsilon \sim \mathcal{N}(0,I)$ is the added noise at the $k$-th diffusion step, and $\epsilon_{\theta}(\cdot)$ is the added predicted noise from the model.
$\mathcal{T}_k = \sqrt{\bar{\alpha}_k}\mathcal{T}_0 +
    \sqrt{1-\bar{\alpha}_k}\epsilon$ is the noisy action chunk perturbed from $\mathcal{T}_0$, where $\bar{\alpha}_k$ is the cumulative product of the noise schedule.
% $w_{\tau}$ is a weighting term to balance the loss across different diffusion steps.
The auxiliary losses include three terms:
$\mathcal{L}_{\mathrm{wp}}$ is the waypoint prediction MSE loss;
$\mathcal{L}_{\mathrm{occ}}$ is the local occupancy reconstruction loss, for both small and downsampled large patches with binary cross entropy loss;
and $\mathcal{L}_{\mathrm{sdf}}$ is the SDF loss to enhance the collision avoidance performance, defined as 
$    \mathcal{L}_{\mathrm{sdf}} =
    \left[
    \operatorname{ReLU}
    \left(d_{\mathrm{thres}} - d_{\mathrm{sdf}}\right)
    \right]^2$,
where $d_{\mathrm{sdf}}$ is the signed distance from the generated trajectory or discrete waypoints to the nearest obstacle, and $d_{\mathrm{thres}}$ is a safety threshold.

% With the loss defined in Eq.~\eqref{eq:overall_loss}, 
We train Roken following a pre-training and post-training paradigm.
The pre-training stage trains the model with the loss function $\mathcal{L}$ on expert datasets, which alone has shown strong performance in generating collision-free and communication-maintained trajectories. 
However, the model occasionally falls short in escaping from the local minima areas in cluttered environments. 
Therefore, we apply the post-training stage with trajectory-level reinforcement learning to further improve the safety and goal-reaching performance of the model.
Specifically, we adopt the group relative policy optimization (GRPO) algorithm~\cite{shao2024deepseekmath} for critic-free policy optimization. 
During policy rollout, we sample an initial configuration from a random time step of the expert trajectory, and perform parallel generation of multi-robot trajectories, covering both safe and unsafe candidates. The rewards of each candidate are then scored within the group.
Note that we inject the privilege information about the geodesic distance to the reward function to enhance the global planning performance of the model, rather than only focusing on the short-term best candidates.

% \textbf{Privilege information-driven trajectory-level reinforcement learning}
% To further enhance safety constraints and the global planning performance of the model, we apply trajectory-level reinforcement learning with privilege information after the imitation learning stage.
% The privilege information about the geodesic distance to the goal and the signed distance to obstacles can be obtained from the dataset and used to construct reward functions for reinforcement learning.
% With this, we penalize the trajectories that hit obstacles or trapped into local minima area (where the geodesic distance to the goal does not decrease), and reward the trajectories that make progress towards the goal.
% We adopt the group relative policy optimization (GRPO) algorithm for multi-robot trajectory-level policy optimization, which is a multi-agent extension of proximal policy optimization (PPO) that optimizes a shared policy with group-level trajectories~\cite{liang_GroupRelativePolicyOptimization_2024}.
% The algorithm alleviate the training of a critic network, and allows better sample efficiency.
% Specifically, we rollout the model in the training environments, where the initial configuration is sampled at a random time step of the expert trajectory, and perform parallel generation of multi-robot trajectories. 
% The reward of each team trajectories are then scored within the group.
% During online policy training, these trajectories can be mixed as their scores have been normalized within the group, which can stabilize the training and improve the sample efficiency.

\section{Experimental Results}

We evaluate Roken around capability, generalizability, and scalability.
Specifically, we evaluate 
(1) whether a unified DiT model can learn to generate coordinated multi-robot trajectories that satisfy safety and communication constraints; 
(2) what is the effect of auxiliary tasks compared to only using the denoising objectives; 
(3) whether Roken can scale with varying team sizes and generalize to unseen and out-of-distribution maps.
Our model is implemented in PyTorch and trained on an NVIDIA A6000 GPU.
The model contains about 11.9 million parameters, with 8 transformer decoder layers and a hidden dimension of 512. 
% It has 8 transformer decoder layers, 8 attention heads, and a hidden dimension of 512. 
We pad the robot tokens to a maximum of 10. 
The occupancy map is padded to a size of 140$\times$140, downsampled by a ResNet~\cite{he2016deep} encoder and then divided into discrete 10$\times$10 patches.
The AdamW optimizer~\cite{loshchilovdecoupled} is used with a learning rate of 5e-4.

% Our model is implemented in PyTorch and trained on an NVIDIA A6000 GPUs with a batch size of 128.
% The model contains about 11.9 million parameters. 
% It has 8 transformer decoder layers, 8 attention heads, and a hidden dimension of 512. 
% We pad the robot tokens to a maximum of 10 robots. 
% The occupancy map is padded to a size of 140$\times$140, downsampled by a ResNet~\cite{he2016deep} encoder to half its original resolution, and then divided into discrete 10$\times$10 patches.
% The AdamW optimizer \red{Here: reference} is used with a learning rate of 5e-4.

\textbf{Baseline methods:} We compare Roken with three baselines that represent different paradigms for multi-robot connectivity maintenance. \textbf{Laplacian}~\cite{bai2026lineofsightconstrainedmultirobotmaplessnavigation} is a continuous gradient field-based approach based on graph-Laplacian for connectivity maintenance. 
\textbf{SGG}~\cite{shi_CommunicationAwareMultirobot_2021} formulates multi-robot coordination as a discrete topology optimization problem with continuous refinement.
% It first searches for the next minimal connected topology based on evenly sampled candidates around each robot's current position, followed with a continus optimization process to refine the next positions to minimize the motion cost and ensure safety constraints. 
\textbf{GNN}~\cite{tang_DecentralizedCommunicationMaintained_2024} trains a Graph Neural Network (GNN)-based model to score candidate future topology graphs of robots, and the next position for each robot is selected sequentially based on the scores.
% Each robot constructs a set of candidate graphs based on its discrete candidate positions along with other robots' current positions, and the GNN model scores these candidate graphs based on their connectivity and collision avoidance properties. 

\textbf{Evaluation metrics:} 
We evaluate the performance of different methods on 1000 unseen environments with varying layouts and initial configurations. 
Based on the problem formulation, we define five metrics to evaluate the performance of different methods: obstacle collision ratio (Obs-Collision), inter-robot collision ratio (Inter-Collision), team connectivity ratio, goal reach ratio (Reach), and distance efficiency (Length).
The distance efficiency is measured by the average trajectory length of all robots in successful episodes, normalized by the discrete A* path length.
We summarize the performance with Full Success, where a successful episode requires the team to reach the target while \emph{always} satisfying both safety and communication constraints.
Therefore, the full success ratio is a stringent and comprehensive metric, along with the hard collision avoidance metrics.
The connectivity ratio or reach ratio is less reliable because the model can generate invalid trajectories that violate safety constraints but still get high ratios.

\subsection{Performance Evaluation on Distribution Learning}

\begin{table}[t]
\centering
\begin{minipage}[t]{.7\linewidth}
\vspace{0pt}
\caption{Evaluation results on unseen environments with four robots.}
\label{tab:v19_ablation}
\centering
\scriptsize
\setlength{\tabcolsep}{1.2pt}
\resizebox{\linewidth}{!}{%
\begin{tabular}{lcccccccccc}
\hline
Model & $\mathcal{L}_{\mathrm{traj}}$ & $\mathcal{L}_{\mathrm{occ}}$ & $\mathcal{L}_{\mathrm{wp}}$ & $\mathcal{L}_{\mathrm{sdf}}$ & \shortstack{Full\\success $\uparrow$} & \shortstack{Obs-\\Colli. $\downarrow$} & \shortstack{Inter-\\Colli. $\downarrow$} & Connect. $\uparrow$ & Reach $\uparrow$ & Length $\downarrow$ \\
\hline
SGG~\cite{shi_CommunicationAwareMultirobot_2021} & \multicolumn{4}{c}{--} & 0.479 & \textbf{0.000} & \underline{0.028} & 0.505 & 0.976 & \textbf{0.492} \\
GNN~\cite{tang_DecentralizedCommunicationMaintained_2024} & \multicolumn{4}{c}{--} & 0.327 & 0.329 & \textbf{0.000} & 0.414 & \underline{0.999} & 2.174 \\
Laplacian~\cite{bai2026lineofsightconstrainedmultirobotmaplessnavigation} & \multicolumn{4}{c}{--} & 0.741 & \underline{0.098} & 0.053 & \underline{0.949} & 0.784 & 1.178 \\
\hline
Roken$_\text{four}$ & $\checkmark$ & $\checkmark$ & $\checkmark$ & $\checkmark$ & 0.765 & 0.145 & 0.081 & 0.873 & 0.923 & 0.623 \\
Roken\textsubscript{mixed} & $\checkmark$ & $\checkmark$ & $\checkmark$ & $\checkmark$ & \underline{0.779} & 0.112 & 0.098 & 0.899 & 0.915 & 0.629 \\
Roken\textsubscript{RL} & $\checkmark$ & $\checkmark$ & $\checkmark$ & $\checkmark$ & \textbf{0.790} & 0.127 & 0.089 & 0.920 & 0.947 & 0.622 \\
\hline
\multirow{4}{*}{\shortstack{Roken$_\text{four}$\\Ablations}} & $\checkmark$ & $\times$ & $\times$ & $\times$ & 0.262 & 0.707 & 0.110 & \textbf{0.997} & \textbf{1.000} & \underline{0.523} \\
 & $\checkmark$ & $\checkmark$ & $\times$ & $\times$ & 0.724 & 0.179 & 0.088 & 0.844 & 0.894 & 0.631 \\
 & $\checkmark$ & $\checkmark$ & $\checkmark$ & $\times$ & 0.698 & 0.225 & 0.101 & 0.874 & 0.910 & 0.619 \\
 & $\checkmark$ & $\checkmark$ & $\checkmark$ & $\checkmark$ & 0.719 & 0.164 & 0.122 & 0.865 & 0.909 & 0.628 \\
\hline
\end{tabular}
}
\vspace{2pt}
\parbox{\linewidth}{\footnotesize \textit{Note:} All Roken$_\text{four}$ ablation models are trained for 200 epochs on the same four-robot datasets. All methods are evaluated on 1000 unseen environments with five independent runs, which in total involve 5000 evaluation episodes. Bold: best performance; Underline: second-best performance.}
\end{minipage}\hfill
\begin{minipage}[t]{.27\linewidth}
\vspace{10pt}
\centering
\includegraphics[width=\linewidth]{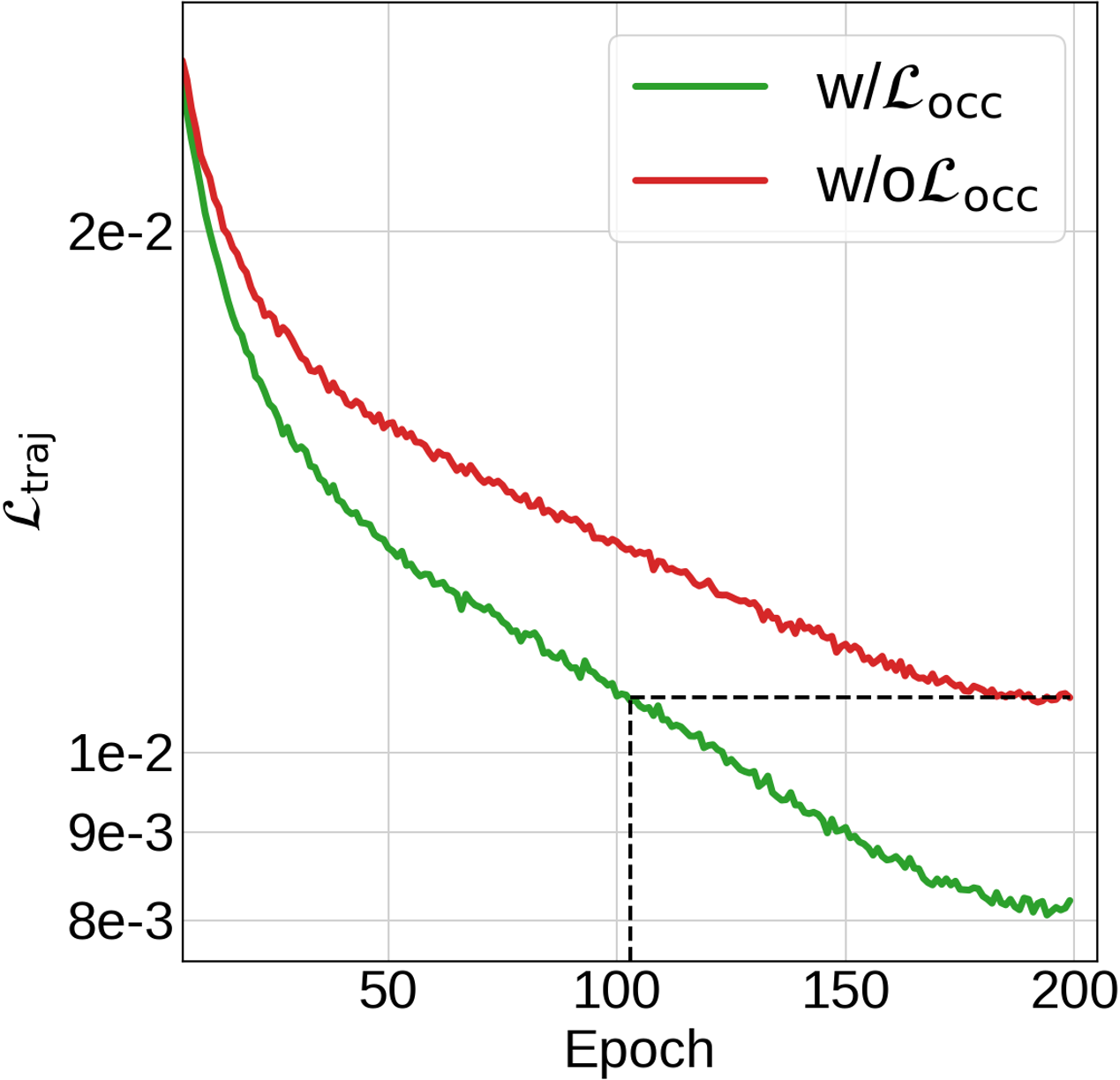}
\vspace{-8pt}
\captionof{figure}{Training loss of $\mathcal{L}_{\mathrm{traj}}$ w/ and w/o $\mathcal{L}_{\mathrm{occ}}$ in Roken$_\text{four}$ ablations.}
\label{fig:loss_noise}
\end{minipage}
\vspace{-20pt}
\end{table}

\textbf{Comparison with baseline methods:} We first compare the performance of Roken with the baseline methods on the unseen test environments, as shown in Table~\ref{tab:v19_ablation}.
It is worth noting that SGG, GNN, and Laplacian use the global optimal A* path for the leader robot, while Roken does not. 
Despite that, both of our models Roken$_{\text{four}}$ (trained on four-robot dataset) and Roken$_{\text{mixed}}$ (trained on mixed dataset with two to eight robots) outperform the baseline methods.
Roken$_{\text{mixed}}$ performs slightly better than Roken$_{\text{four}}$ in terms of full success ratio, which indicates that absorbing training data with diverse team sizes can help to learn more generalized coordination behaviors.
%  with a success rate of 0.779, which means Roken learns the successful strategies from the dataset to avoid being trapped in local minima areas, and generates more distance efficient trajectories.
The post-trained model Roken$_{\text{RL}}$ further improves the full success rate to 0.790, with better reach ratio and connectivity ratio, demonstrating the successful enhancement of the global planning performance of the model.
In contrast, although the compared baseline methods follow safety constraints better than Roken, they fail to maintain connectivity and reach the goal at the same time, leading to much lower full success ratios. 
% While both Laplacian and SGG can follow the safety constraints well, SGG fails to maintain connectivity during robot navigation as robots frequently get trapped in local minima areas and fail to follow the leader robot.
The Laplacian method achieves a much higher full success rate of 0.741, but it suffers from a lower reach ratio because robots are frequently stuck in conflicting potential fields, which also results in longer zig-zag trajectories.
% Moreover, the averaged trajectory length is much longer due to its zig-zagging behavior induced by the changing gradient field.

\textbf{Ablation studies:}
We train a series of Roken$_{\text{four}}$ ablation models to investigate the contribution of each auxiliary loss.
As shown in Tab.~\ref{tab:v19_ablation}, with only $\mathcal{L}_{\mathrm{traj}}$, a vanilla DiT model is difficult to learn the map-conditioned trajectory distribution, with the highest obstacle collision ratio.
% A vanilla DiT model is therefore incapable of learning the complex map-conditioned trajectory distribution.
The largest improvement happens after adding $\mathcal{L}_{\mathrm{occ}}$, which significantly improves the full success ratio from 0.262 to 0.724.
Besides that, $\mathcal{L}_{\mathrm{occ}}$ also helps to double the training speed with a much lower $\mathcal{L}_{\mathrm{traj}}$, as shown in Fig.~\ref{fig:loss_noise}.
This supports our analysis on the conditional distribution learning with auxiliary occupancy reconstruction loss.
The $\mathcal{L}_{\mathrm{wp}}$ improves the reach ratio by avoiding local minima areas through long-term waypoint prediction, but it competes with the trajectory tasks and therefore leads to a slight drop of the full success rate. 
$\mathcal{L}_{\mathrm{sdf}}$ additionally helps to improve the collision avoidance performance, achieving the lowest obstacle collision ratio of 0.164 in all ablation models.

% \red{Qualitative rollout results show that Roken can generate coordinated trajectories in unseen environments. The generated motions exhibit three behaviors observed in the preliminary report: robots tend to preserve communication, inactive robots move lazily when possible, and the team adapts its minimum topology as the leader moves through the environment. In contrast to reactive controllers that only correct constraint violations after they appear, Roken predicts a short-horizon team trajectory in one denoising rollout and replans periodically during execution.}

\subsection{Scalability Evaluation with Varying Team Sizes}

\begin{figure}[t]
    \centering
    \includegraphics[width=\linewidth]{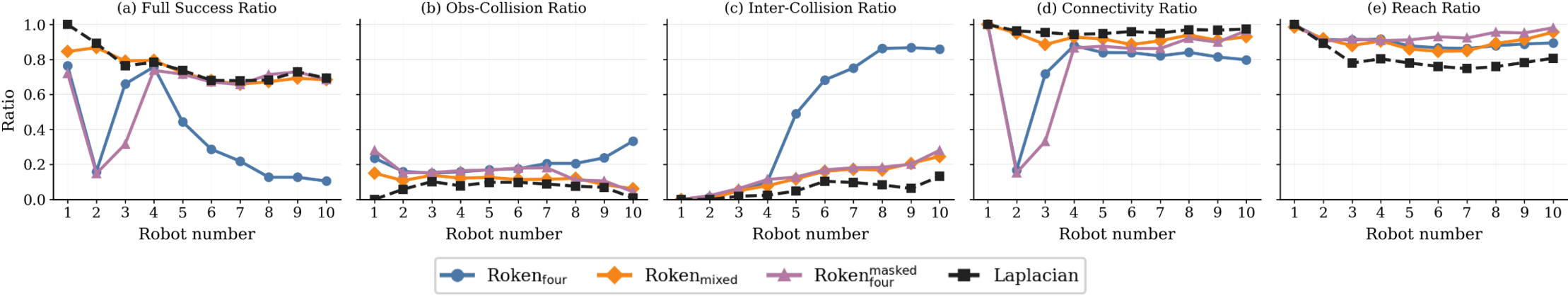}
    \caption{Scalability evaluation from one to ten robots for three variants of Roken and the baseline method. We rolled out 500 episodes in 500 unseen maps for each team size and model variant.}
    \label{fig:scalability}
    \vspace{-10pt}
\end{figure}

Roken is designed to be scalable to different team sizes, in both training and inference.
We compare three variants of Roken trained on different recipes of datasets:
Roken$_{\text{four}}$, Roken$_{\text{four}}^{\text{masked}}$, and Roken$_{\text{mixed}}$, where Roken$_{\text{four}}^{\text{masked}}$ randomly masks robot tokens during training for conditional generation.
%  is trained on the same four-robot dataset but with random masking of robot tokens during training, which enables conditional generation and therefore can be applied to different team sizes during inference by clustered generation; 
% Roken$_{\text{mixed}}$ is trained on a mixed dataset that includes trajectories with two to eight robots.
We also compared with the baseline Laplacian method~\cite{bai2026lineofsightconstrainedmultirobotmaplessnavigation}.
The results are reported in Fig.~\ref{fig:scalability}.

Overall, the scalability performance of Roken$_{\text{mixed}}$, Roken$_{\text{four}}^{\text{masked}}$, and the baseline Laplacian method~\cite{bai2026lineofsightconstrainedmultirobotmaplessnavigation} are similar, especially when the team size is large.
However, it is worth noting that Laplacian uses the global optimal A* path for the leader robot as the reference trajectory, which is not available for Roken.
Specifically, Roken$_{\text{mixed}}$ shows good scalability performance across different team sizes, with a more stable full-success ratio as the team size increases, even though the training data distribution is heavily imbalanced towards the four-robot case, as shown in Fig.~\ref{fig:dataset}.
This verifies our assumption that by the robot-token design, Roken learns generalizable coordination behaviors across diverse team sizes.

Without masked training, the Roken$_{\text{four}}$ model only scales to 3 or 5 robots with moderate success ratios, but fails to scale to more robots. 
In contrast, the masked training of Roken$_{\text{four}}^{\text{masked}}$ provides conditional planning capability to the model, thereby can scale to much larger team sizes with clustered rollout, as shown in Fig.~\ref{fig:scalability}.
The performance of clustered rollout is even similar to the direct rollout of Roken$_{\text{mixed}}$ trained with mixed team sizes.
Therefore, we claim that the masked training and clustered rollout provides a scalable way to either directly apply the pre-trained Roken to larger team sizes without additional training, or to generate expert trajectories of larger teams for further pre-training of Roken.

% All three variants show high full success ratios for single-robot planning (although we never train them on single-robot datasets), and high goal-reaching ratios for all team sizes, indicating that the model learns strong goal-oriented behaviors especially for the leader robot.
% Roken$_{\text{four}}$ shows some generalization to three or five robots, but fails to capture the coordinated movements as team size increases, leading to frequent inter-robot collisions.
% With clustered conditional generation, Roken$_{\text{four}}^{\text{masked}}$ can achieve better success ratio for larger team sizes, with good collision avoidance performance. 
% However, it also introduces disturbances for the model to learn the connectivity maintenance behaviors, as indicated by the significant drop of connectivity ratio with larger team sizes.

% \begin{figure*}[t]
%     \centering
%     \includegraphics[width=\textwidth]{images/fig_scalability.pdf}
%     \vspace{-10pt}
%     \caption{}
%     \label{fig:scalability}
%     \vspace{-10pt}
% \end{figure*}

% \subsection{Details of Scalability Experiments}

% Note that the scalability performance of Roken$_{\text{four}}^{\text{masked}}$ is better than the results in Fig.~\ref{fig:scalability} in the main paper, after we fix a bug in previous code.
% The model Roken$_{\text{four}}^{\text{masked}}$ remains unchanged as in the main paper.

% The updated results are shown in Fig.~\ref{fig:scalability_more}.

\subsection{Generalizability and Real-world Deployment}
\label{sec:generalizability}

\begin{wraptable}{r}{.32\linewidth}
    \vspace{-20pt}
    \centering
    \small
    \setlength{\tabcolsep}{5pt}
    \caption{Success ratios under different sensing ranges.}
    \vspace{-5pt}
    \begin{tabular}{lc}
    \hline
    Sensing range & Full success \\
    \hline
    Full map & 0.836 \\
    20 & 0.832 \\
    10 & 0.782 \\
    5 & 0.600 \\
    \hline
    \end{tabular}
    \label{tab:sensing_range_full_success}
    \vspace{-20pt}
\end{wraptable}

This section evaluates the generalizability of Roken from global maps to local and out-of-distribution maps.
We conduct zero-shot evaluation of Roken$_{\text{RL}}$ by inputting only partially observed maps around robots, which is more realistic for real-world applications.
The full success ratios in 500 episodes with unseen maps are reported in Table~\ref{tab:sensing_range_full_success}, with different sensing ranges defined in the pixel space.
When the sensing range is large, the model maintains a similar success ratio as the global map input, verifying its map query capability.
However, as the sensing range decreases to be very small, the generated trajectories are more likely to get stuck before realizing the map structure, which indicates the limitation of our model in escaping from local minima areas with only short-term observation.
We also deploy Roken on a real-world multi-robot system with various numbers of omnidirectional robots, as shown in Fig.~\ref{fig:intro_capabilities}.
We manually design cluttered environments that are different from the training maps.
Despite that, Roken can still generate coordinated trajectories that maintain connectivity and avoid obstacles, which verifies the real-world applicability of the model.

% \begin{figure}[t]
%     \centering
%     \includegraphics[width=\linewidth]{images/fig_results.pdf}
%     \vspace{-5pt}
%     \caption{Simulation rollout (left) and real-world deployment (right) of Roken.}
%     \label{fig:realworld}
%     \vspace{-15pt}
% \end{figure}

\section{Conclusion}

This work studies the connectivity-constrained multi-robot trajectory generation problem. 
We proposed Roken, a unified DiT model that represents robots as tokens, learns and generates coordinated multi-robot trajectories in a feed-forward manner. 
We propose multi-scale spatial-temporal self-supervision to enhance the conditional distribution learning of the model, and further improve the safety and long-horizon planning capabilities of the model through trajectory-level reinforcement learning. 
Roken demonstrates strong capability in generating coordinated trajectories that satisfy safety and communication constraints, outperforming baseline methods used to collect training datasets. 
After training on a mixed dataset with varying team sizes, Roken shows good scalability to different team sizes with stable success rates.
We also demonstrate the model's ability to generalize to partially known and out-of-distribution maps, with successful real-world deployment.

\section{Limitations}
As mentioned in Sec.~\ref{sec:generalizability}, Roken falls short in generating the escaping trajectories once it gets stuck in local minima areas, mainly because the training data lacks sufficient examples of such scenarios.
% Although Roken has shown better planning and coordination performance than baselines, it still show some failure cases in local minima areas, where the model fails to escape from local minima and reach the goal.
Moreover, the current local occupancy reconstruction tasks only require the model to reconstruct a pre-defined small and large patches, which limits the robot tokens to acquire the global horizon.
Finally, despite the improved obstacle avoidance performance of the model, a safety shield may still be needed to ensure the safety in real-world applications.

\newpage

% The current model has three main limitations. First, long-range planning is still difficult when a short action chunk is insufficient to reveal the correct homotopy class. Second, communication maintenance sometimes requires smarter coordination than local trajectory denoising can provide, especially around obstacles where a relay robot must move away from its own shortest path. Third, scalability through clustered inference is promising but not yet a formal guarantee; cluster construction and consistency across overlaps need further study. Future work will investigate safety shields, reinforcement learning fine-tuning, and simulation or real-world demonstrations.

% The acknowledgments are automatically included only in the final and preprint versions of the paper.
% \acknowledgments{\red{TODO: add final acknowledgments, funding statement, and collaborator thanks if needed.}}

% no \bibliographystyle is required, since the CoRL style is automatically used.
\bibliography{example}

\appendix
\newpage

\section{Hyperparameters}

\textbf{Task specifications:}
\begin{table}[h]
    \centering
    \caption{Task constraint parameters used in evaluation.}
    \label{tab:task_specifications}
    \begin{tabular}{lcl}
        \hline
        Parameter & Value & Description \\
        \hline
        $R$ & 15.0 & Maximum communication radius between two robots \\
        $d_{\mathrm{coll}}^{\min}$ & 2.0 & Minimum allowed distance between two robots \\
        $d_{\mathrm{obs}}^{\min}$ & 1.0 & Minimum allowed distance from a robot to obstacles \\
        $\epsilon_g$ & 1.0 & Goal-reaching threshold distance \\
        \hline
    \end{tabular}
    \parbox{\linewidth}{\footnotesize \textit{Note:} All values are in the pixel space, where 1 pixel corresponds to 1 meter in the real world.}
\end{table}

\textbf{Roken model:}
\begin{table}[h]
    \centering
    \caption{Roken model hyperparameters.}
    \label{tab:model_hyperparameters}
    \begin{tabular}{lc}
        \hline
        Hyperparameter & Value \\
        \hline
        Number of transformer decoder layers & 8 \\
        Number of attention heads & 8 \\
        Hidden dimension & 512 \\
        Maximum number of robot tokens & 10 \\
        Occupancy map size (pixels) & 140$\times$140 \\
        Occupancy patch size (pixels) & 10$\times$10 \\
        Reconstructed local small patch size (pixels) & 7$\times$7 \\
        Reconstructed local large patch size (pixels) & 7$\times$7 \\
        Number of predicted waypoints & 5 \\
        Waypoints sampling interval (meters) & 8 \\
        Action chunk prediction horizon & 16 \\
        Action maximum step size (meters) & 0.5 \\
        Trajectory sampling interval (steps) & 4 \\
        Learning rate & 5e-4 \\
        Batch size & 128 \\
        $\lambda_1$ (trajectory loss weight) & 1.0 \\
        $\lambda_2$ (waypoint loss weight) & 1.0 \\
        $\lambda_3$ (occupancy loss weight) & 0.1 \\
        $\lambda_4$ (SDF loss weight) & 1.0 \\
        $d_{\mathrm{thres}}$ (SDF safety threshold, pixels) & 4.0 \\
        \hline
    \end{tabular}
\end{table}

\newpage
\section{Dataset Collection Pipeline}

\begin{figure}[h]
    \centering
    \includegraphics[width=\linewidth]{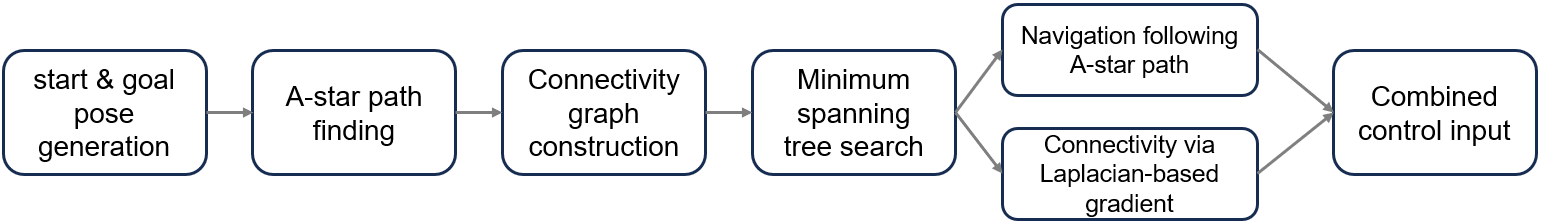}
    \caption{Pipeline for dataset collection.}
    \label{fig:data_generation}
\end{figure}

This section provides detailed description of the dataset collection pipeline, which is based on a classical gradient field-based method adopted from~\cite{nestmeyer_DecentralizedSimultaneous_2017} and~\cite{bai2026lineofsightconstrainedmultirobotmaplessnavigation}.
The detailed process is illustrated in Fig.~\ref{fig:data_generation}.
Specifically, a communication graph $\mathcal{G}_t$ is first constructed as defined in Sec.~\ref{sec:communication_graph}.
The edge weight $w_{ij}$ for each edge $(i, j)$ is defined as the product of the potential function values for different inter-robot constraints, as defined in~\cite{bai2026lineofsightconstrainedmultirobotmaplessnavigation}. 
The difference is that in this paper we do not consider the line-of-sight constraint, and therefore the edge weight is only determined by the communication distance between two robots and the distance to obstacles.
Given the weighted communication graph $\mathcal{G}_t$, a minimum spanning tree $\mathcal{T}_{\text{G}}$ is then searched to determine the minimum topology for connectivity maintenance, where the effort for maintaining the connectivity while navigating to the goal point is minimized. 
With $\mathcal{T}_{\text{G}}$ and its related graph Laplacian matrix $\mathbf{L}_{\mathcal{T}_{\text{G}}}$, the gradient field for maintaining team connectivity is derived as
\begin{equation}
    \boldsymbol{u}_i^c
= - \frac{\partial V^\lambda(\lambda_2)}{\partial \lambda_2}
\cdot \frac{\partial \lambda_2}{\partial p_t^i}
= - \frac{\partial V^\lambda(\lambda_2)}{\partial \lambda_2}
\cdot \sum_{j \in \mathcal{N}_i}
\frac{\partial A_{ij}}{\partial p_t^i}
\left( v_{2i} - v_{2j} \right)^2 ,
\end{equation}
where $V^\lambda(\lambda_2)$ is a potential function of the second smallest eigenvalue $\lambda_2$ of $\mathbf{L}_{\mathcal{T}_{\text{G}}}$ defined as $V^\lambda(\lambda_2)
= \frac{1}{\lambda_2 - \lambda_2^{\min}}$, where $\lambda_2^{\min}$ is the preferred minimum value of $\lambda_2$; $v_{2i}$ is the $i$-th element of the corresponding eigenvector $\boldsymbol{v}_2$; $\mathcal{N}_i$ is the set of neighbors of robot $i$; and $A_{ij}$ is the edge weight between robot $i$ and $j$.

To avoid robots getting stuck when navigating to the goal point, we provide the optimal A* path for the leader robot as a reference trajectory (note that Roken does not have such a reference trajectory, and only relies on the map input for end-to-end trajectory planning). 
The final movement of the leader robot is determined by the combination of the connectivity potential field and the navigation potential field, defined as
\begin{equation}
    \boldsymbol{u}_i
= \lambda_c \cdot \boldsymbol{u}_i^c + \lambda_r \cdot \boldsymbol{u}_i^n,
\end{equation}
where $\boldsymbol{u}_i^n$ is the navigation potential field defined as $\boldsymbol{u}_i^n = \frac{g_t - p_t^i}{\Vert g_t - p_t^i \Vert}$, where $g_t$ is a reference waypoint along the reference A* trajectory for the leader robot.
The two potential fields are normalized and weighted by $\lambda_c$ and $\lambda_r$ to determine the final movement of the leader robot.
For the remaining robots, their movements are determined by the connectivity potential field only, which coordinates them to follow the leader robot while maintaining connectivity.

\begin{figure}[h]
    \centering
    \includegraphics[width=.7\linewidth]{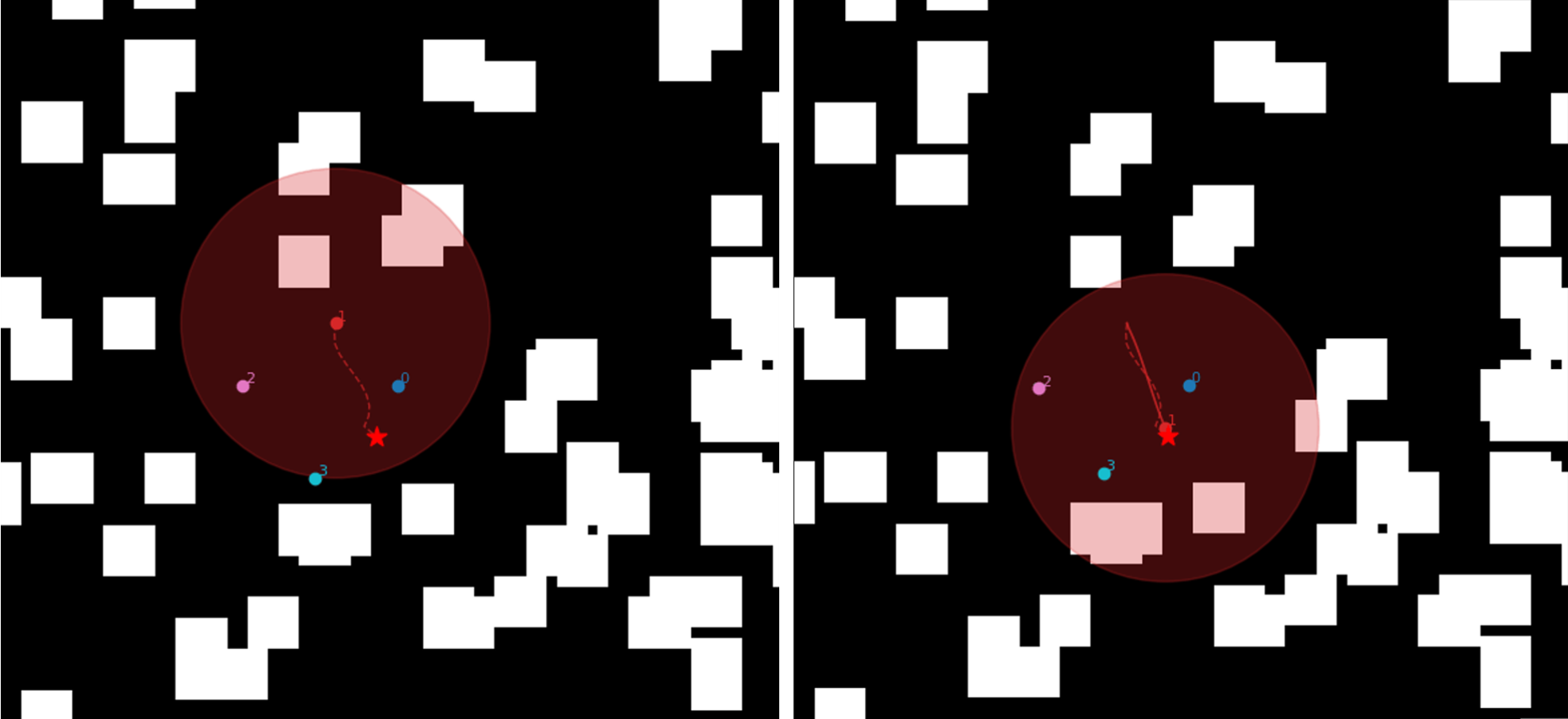}
    \caption{Example episode showing the lazy movement behavior. The leader robot (red) moves towards the goal, while the other robots (blue) remain stationary because they are always in communication range. The dotted line is the smoothed optimal A* path only for visualization purposes.}
    \label{fig:lazy_movement}
\end{figure}

\textbf{Lazy movement:} Because of the minimum spanning tree search module, the final robot trajectories exhibit a \emph{lazy movement} behavior, where robots only move when it is necessary to maintain connectivity, and they tend to stay still when possible. 
This is to avoid unnecessary movements and to make robots more dispersed in the environment with larger simultaneous coverage.
An example episode is shown in Fig.~\ref{fig:lazy_movement}.

\newpage
\section{Loss Functions}

This section provides detailed loss function definitions in Eq.~\eqref{eq:overall_loss}.
Recall that the overall loss function is defined as 
\begin{equation}
\mathcal{L}
=
\lambda_{1}\mathcal{L}_{\mathrm{traj}}
+ \lambda_{2}\mathcal{L}_{\mathrm{wp}}
+ \lambda_{3}\mathcal{L}_{\mathrm{occ}}
+ \lambda_{4}\mathcal{L}_{\mathrm{sdf}}.
\end{equation}
Besides the trajectory denoising loss $\mathcal{L}_{\mathrm{traj}}$, the other three auxiliary losses are defined as follows.
Note that all the three auxiliary losses can be calculated with the self-contained information from the training dataset, without the need of additional data collection or annotation, which we call multi-scale spatial-temporal self-supervision.

\textbf{Waypoint Prediction Loss $\mathcal{L}_{\mathrm{wp}}$.}
The waypoint head predicts $K$ (we take $K=5$) sparse future waypoints for each robot, defined as 
$\hat{L}^i = \{\hat{l}^i_1,\ldots,\hat{l}^i_K\}$,
which are sampled from the expert trajectory in the training dataset at fixed distance intervals. 
If no additional waypoints can be sampled before reaching the goal, the remaining waypoints are set to the goal position.
% The waypoints are 
% \begin{equation}
%     \hat{L}^i = \{\hat{l}^i_1,\ldots,\hat{l}^i_K\}.
% \end{equation}
The waypoint loss is the mean squared error between predicted and target anchors:
\begin{equation}
    \mathcal{L}_{\mathrm{wp}} =
    \frac{1}{NK}\sum_{i=1}^{N}\sum_{r=1}^{K}
    \left\|
    \hat{l}^{i}_{r} - l^{i}_{r}
    \right\|_2^2.
\end{equation}
This auxiliary target improves long-range planning by encouraging robot tokens to encode future progress beyond the short action chunk.

\textbf{Occupancy Reconstruction Loss $\mathcal{L}_{\mathrm{occ}}$.}
We define two local occupancy reconstruction heads that decode two occupancy patches from middle-layer latent vectors at two spatial scales (1$\times$ and 3$\times$ in our implementation).
The sizes of the occupancy patches are set to 7$\times$7.
Therefore, this task guides the model to learn the coarse spatial structure within a large 21$\times$21 area around each robot, and the fine-grained structure within a small 7$\times$7 area.
The large patch is downsampled to the same 7$\times$7 resolution as the small patch to reduce the capacity burden of the latent vectors.
Given the ground-truth occupancy labels $y^q_{hw}$ for each patch at scale $q \in \{\mathrm{small}, \mathrm{large}\}$, where $y^q_{hw} = 1$ if the cell is occupied by an obstacle, and $y^q_{hw} = 0$ otherwise, we supervise the occupancy reconstruction heads with binary cross entropy loss:
\begin{equation}
    \mathcal{L}_{\mathrm{occ}}^{q}
    = -\frac{1}{HW}\sum_{h,w}
    [y^q_{hw}\log\sigma(x^q_{hw})
    + (1-y^q_{hw})\log(1-\sigma(x^q_{hw}))],
\end{equation}
where $\sigma(x) = 1/(1 + e^{-x})$.
The combined occupancy loss is
\begin{equation}
    \mathcal{L}_{\mathrm{occ}}
    =
    \mathcal{L}_{\mathrm{occ}}^{\mathrm{small}}
    +
    \mathcal{L}_{\mathrm{occ}}^{\mathrm{large}}.
\end{equation}

\textbf{Signed Distance Loss.}
To encourage obstacle clearance, we use an SDF penalty on generated trajectories and discrete waypoints.
We pre-construct the SDF map for each training environment, where the value of each cell is the signed distance to the nearest obstacle (positive for free space and negative for occupied space).
With the bilinear interpolation of the SDF map, we can obtain the signed distance for any point in the environment, with continuous gradients.
As introduced in Sec.~\ref{sec:training_objectives}, the SDF loss is defined as
\begingroup
\small
\begin{equation}
    \mathcal{L}_{\mathrm{sdf}} =
    \frac{1}{N(T+K)}\sum_{i=1}^{N}\left(
    \lambda_{\mathrm{sdf}}^{\mathrm{traj}}
    \sum_{t=1}^{T}
    \left[
    \operatorname{ReLU}
    \left(d_{\mathrm{thres}} - d_{\mathrm{sdf}}(p_t^i)\right)
    \right]^2
    +
    \sum_{r=1}^{K}
    \lambda_{\mathrm{sdf}}^{\mathrm{wp}}
    \left[
    \operatorname{ReLU}
    \left(d_{\mathrm{thres}} - d_{\mathrm{sdf}}(\hat{l}_r^i)\right)
    \right]^2
    \right),
\end{equation}
\endgroup
where $d_{\mathrm{sdf}}(p)$ is the signed distance for point $p$ obtained from the SDF map, and $d_{\mathrm{thres}}$ is a pre-defined distance threshold for obstacle clearance.
We set both $\lambda_{\mathrm{sdf}}^{\mathrm{traj}}$ and $\lambda_{\mathrm{sdf}}^{\mathrm{wp}}$ to 1.0 in our implementation.

\newpage
\section{Trajectory-Level Reinforcement Learning}

\textbf{Rollout dataset collection:}
We perform post-training reinforcement learning on top of the supervised pretrained Roken. 
Each RL iteration first constructs an on-policy rollout batch from the initial states sampled from the expert episode mid-states. 
Concretely, we sample an expert episode from the training set and randomly choose an intermediate timestep $t$ when the remaining future trajectory is long enough for rollout evaluation. 
The prompt state is then formed as $c=(\mathcal{H}_{\mathrm{map}}, \mathcal{P}_t, g, \{m_g^i\}_{i=1}^{N})$, where $\mathcal{P}_t=\{p_t^i\}_{i=1}^{N}$ denotes the current team positions and $\{m_g^i\}_{i=1}^{N}$ is the binary goal mask. We also require the sampled multi-robot state to satisfy the same communication and safety constraints as defined in Sec.~\ref{sec:constraints}, so that RL starts from feasible team configurations.

For each prompt $c_q$, we generate a group of $G$ candidate team trajectories from the current Roken mode, which can be denoted as a diffusion policy $\pi_\theta$. All $G$ candidates share the same conditioning input, but use independent diffusion noise:
\[
\mathcal{T}_{q,j} \sim \pi_\theta(\cdot \mid c_q), \qquad j=1,\ldots,G .
\]
In our implementation, $G$ is set to 32 for training rollouts. The diffusion model predicts normalized relative action sequences, which are converted into absolute trajectories by integrating from the current robot positions. These trajectories are then transformed back to the environment coordinate frame before reward computation. This produces a rollout batch of grouped candidates
\[
\mathcal{B}
=
\{(c_q, \mathcal{T}_{q,1}, \ldots, \mathcal{T}_{q,G})\}_{q=1}^{P},
\]
where $P$ is the number of prompts collected in one RL rollout. 

Randomly sampling mid-states from the expert trajectories may result in an imbalanced rollout dataset, as most of the sampled states are relatively easy for the pretrained policy by directly navigating to the goal.
To emphasize difficult states, we optionally mix in hard prompts with a pre-defined probability.
The hard prompts are selected from mid-states where the pretrained policy previously generated invalid or low-quality candidates.

\textbf{Reward function with privilege information:}
Each sampled team trajectory $\mathcal{T}_{q, j}$ is evaluated by a privileged reward function with additional simulator-side information, including the signed-distance field, geodesic distance to the goal, collision checks, communication connectivity checks, and the deviation from the optimal A* path. These privileged signals are used only during RL training and are not required at deployment.
For a candidate team trajectory $\mathcal{T}$, the cost function is defined as
\[
C(\mathcal{T})
=
C_{\mathrm{obs}}(\mathcal{T})
+
C_{\mathrm{inter}}(\mathcal{T})
+
C_{\mathrm{conn}}(\mathcal{T})
+
C_{\mathrm{goal}}(\mathcal{T})
+
C_{\mathrm{wp}}(\mathcal{T}),
\]
where the first three terms capture the hard constraints of the task: $C_{\mathrm{obs}}$ penalizes obstacle collision by counting the number of collisions with soft margin; $C_{\mathrm{inter}}$ penalizes robot-robot collisions and soft minimum-distance violations using $\|p_t^i-p_t^j\|_2$; $C_{\mathrm{conn}}$ penalizes communication connectivity failures through $\mathcal{G}_t$ and $\lambda_2(\mathcal{L}_{\mathcal{G}_t})$, by counting the number of disconnected steps.
If above hard constraints are violated, a large cost $C_{\text{max}}$ is assigned to the trajectory to reduce its generation probability.

Besides that, we also inject privileged signals to encourage better planning performance:
$C_{\mathrm{goal}}$ measures leader's final distance to the goal $g$ based on geodesic distance; and $C_{\mathrm{wp}}$ penalizes deviation from the ground truth waypoints sampled from the optimal A* path. 
Note that $C_{\mathrm{wp}}$ is important for improving the long-horizon planning performance of the model, because it is insufficient to score candidate short action chunks to check whether it avoids local minima areas, while the long-horizon waypoints can provide larger horizon for candidate scoring.
We also add a reach bonus once the leader robot reaches the goal to encourage task completion.
The scalar reward is then obtained by clipping and rescaling the total cost:
\[
r(\mathcal{T})
=
-\frac{\min(C(\mathcal{T}), C_{\max})}{s}.
\]
Here $s$ controls the reward magnitude. Thus, safer trajectories with better leader goal progress receive larger rewards, while collision, disconnection, and obstacle violations receive lower rewards.

\textbf{Training settings:}
We use a group-relative policy optimization objective following~\cite{shao2024deepseekmath}. 
Since all $G$ candidates in one group are sampled from the same prompt, their rewards can be compared directly without training a value function. For prompt $q$, let $r_{q,j}=r(\mathcal{T}_{q,j})$ be the reward of candidate $j$. We compute the normalized group-relative advantage
\[
A_{q,j}
=
\frac{
r_{q,j} - \mathrm{mean}_{l=1}^{G}(r_{q,l})
}{
\mathrm{std}_{l=1}^{G}(r_{q,l}) + \epsilon
}.
\]
If the reward standard deviation of a group is too small, all advantages in that group are set to zero, because the group does not provide a useful ranking signal.

During rollout generation, we store the stochastic DDPM reverse-process transitions for each sampled team trajectory using the current policy $\pi_{\theta_{\text{old}}}$. Specifically, for every denoising step, we save the noisy action chunk $\mathcal{T}_k$, the sampled next action chunk $\mathcal{T}_{k-1}$, the timestep $k$, and the behavior-policy log probability
\[
\log \pi_{\theta_{\mathrm{old}}}(\mathcal{T}_{k-1}^{q,j} \mid \mathcal{T}_k^{q,j}, c_q, k).
\]
Each denoising step can be viewed as a decision-making process by sampling the next action chunk $\mathcal{T}_{k-1}$ from a multi-variate Gaussian distribution, denoted as
\begin{equation}
\mathcal{T}_{k-1}^{q,j}
\sim
\pi_{\theta_{\text{old}}}(\cdot \mid \mathcal{T}_k^{q,j}, c_q, k) 
= \mathcal{N}(\mu_{\theta_{\text{old}}}(\mathcal{T}_k^{q,j}, c_q, k), \sigma^{2}_k I).
\end{equation}
The variance $\sigma^{2}_k$ is determined by the noise schedule and is fixed during training, while the mean is determined by the predicted noise $\epsilon_{\theta_{\text{old}}}(\mathcal{T}_k^{q,j}, c_q, k)$ as
\begin{equation}
\mu_{\theta_{\text{old}}}(\mathcal{T}_k^{q,j}, c_q, k)
=
\frac{1}{\sqrt{\alpha_k}}\left(
\mathcal{T}_k^{q,j} - \frac{1-\alpha_k}{\sqrt{1-\bar{\alpha}_k}}\epsilon_{\theta_{\text{old}}}(\mathcal{T}_k^{q,j}, c_q, k)
\right),
\end{equation}
where $\alpha_k$ and $\bar{\alpha}_k$ are the noise schedule parameters. 
% The log probability of the final generated trajectory $\mathcal{T}_0$ can be computed by summing the log probabilities of all the denoising transitions, denoted as
% \begin{equation}
% \log \pi_{\theta_{\mathrm{old}}}(\mathcal{T}_0^{q,j} \mid c_q)
% =
% \sum_{k=1}^{K}
% \log \pi_{\theta_{\mathrm{old}}}(\mathcal{T}_{k-1}^{q,j} \mid \mathcal{T}_k^{q,j}, c_q, k),
% \end{equation}
% where $\mathcal{T}_K^{q,j}\sim \mathcal{N}(0, I)$ is the initial pure noise input for the reverse process.

% The log probability of the sampled next action chunk $\mathcal{T}_{k-1}^{q,j}$ under the current policy can then be computed.

During policy optimization, we recompute the log probability of the same denoising transition under the current policy $\pi_\theta$ and form the likelihood ratio
\[
\rho_{k,q,j}
=
\exp\left(
\log \pi_\theta(\mathcal{T}_{k-1}^{q,j} \mid \mathcal{T}_k^{q,j}, c_q, k)
-
\log \pi_{\theta_{\mathrm{old}}}(\mathcal{T}_{k-1}^{q,j} \mid \mathcal{T}_k^{q,j}, c_q, k)
\right).
\]
The policy is updated with a clipped PPO-style objective:
\[
\mathcal{L}_{\mathrm{GRPO}}
=
-
\mathbb{E}_{k,q,j}
\left[
\min
\left(
\rho_{k,q,j} A_{q,j},
\mathrm{clip}(\rho_{k,q,j}, 1-\epsilon_{\mathrm{clip}}, 1+\epsilon_{\mathrm{clip}}) A_{q,j}
\right)
\right].
\]
To prevent the RL update from destroying the supervised diffusion prior, we keep a frozen reference model $\pi_{\mathrm{ref}}$, initialized from the supervised checkpoint of Roken, and add an analytic Gaussian KL penalty at every denoising transition:
\[
\mathcal{L}
=
\mathcal{L}_{\mathrm{GRPO}}
+
\beta
\mathbb{E}_{k,q,j}
\left[
D_{\mathrm{KL}}
\left(
\pi_\theta(\cdot \mid \mathcal{T}_k^{q,j}, c_q, k)
\;\|\;
\pi_{\mathrm{ref}}(\cdot \mid \mathcal{T}_k^{q,j}, c_q, k)
\right)
\right].
\]
After collecting one rollout buffer, we perform minibatch updates for a small number of epochs, then discard the buffer and sample a new on-policy batch from the updated diffusion policy. This procedure keeps the RL data on-policy while using the frozen reference model to maintain stable multi-robot trajectory generation following the supervised prior.

\textbf{RL post-training hyperparameters:}
\begin{table}
    \centering
    \caption{RL post-training hyperparameters.}
    \begin{tabular}{lc}
        \hline
        Hyperparameter & Value \\
        \hline
        Group size $G$ & 32 \\
        PPO clip threshold $\epsilon_{\mathrm{clip}}$ & 0.1 \\
        KL coefficient $\beta$ & 0.2 \\
        Number of rollout iterations & 200 \\
        Rollout buffer size & 500$\times$32 \\
        Learning rate & 1e-6 \\
        AdamW weight decay & 0.01 \\
        Gradient clipping & 0.5 \\
        \hline
    \end{tabular}
\end{table}

\begin{table}
    \centering
    \caption{Reward cost function parameters.}
    \begin{tabular}{lc}
        \hline
        Cost term & Segment weight \\
        \hline
        Obstacle collision $C_{\mathrm{obs}}$ & 30.0  \\
        Inter-robot collision $C_{\mathrm{inter}}$ & 30.0  \\
        Communication disconnection $C_{\mathrm{conn}}$ & 30.0 \\
        Leader goal progress $C_{\mathrm{goal}}$ &  30.0 \\
        Leader waypoint deviation $C_{\mathrm{wp}}$ & 50.0  \\
        Reach bonus &50.0\\
        Cost clipping $C_{\max}$ & 10000 \\
        Reward scale $s$ & 20 \\
        \hline
    \end{tabular}
    % \parbox{\linewidth}{\footnotesize \textit{Note:} The prefix segment refers to the part of the trajectory before the current prompt state, while the future segment refers to the part of the trajectory after the current prompt state. The leader goal progress term is focused on the leader robot, while the other terms are evaluated for the full team.}
\end{table}

\newpage

\section{Supplementary Experiments Results}

\subsection{Ablations on Relative / Absolute Trajectory Representation}

We investigate the effect of different trajectory representations on the performance of Roken, including the trajectory represented with normalized absolute positions, and a relative representation with relative motion between consecutive time steps.

\begin{figure}[H]
    \centering
    \includegraphics[width=\linewidth]{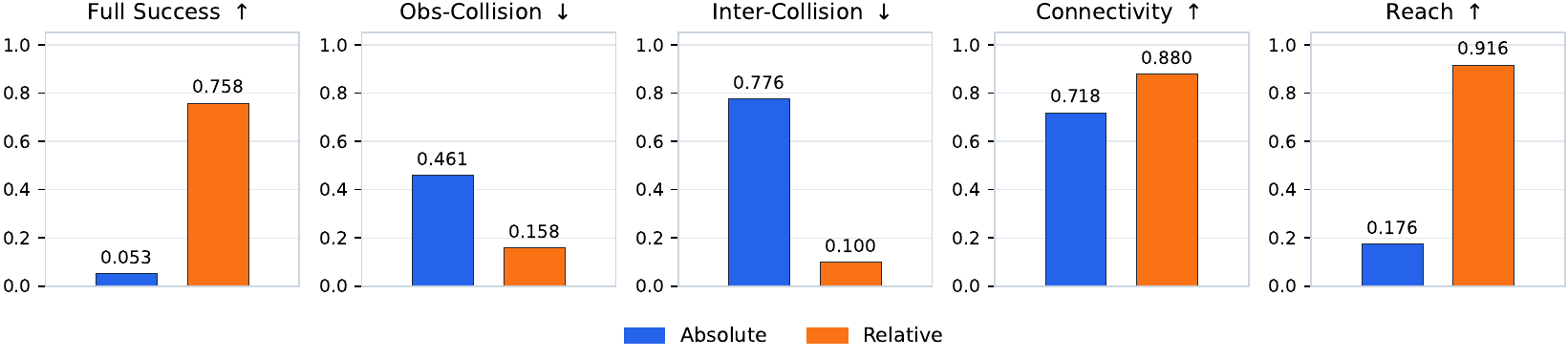}
    \vspace{-8pt}
    \caption{Ablation on absolute and relative trajectory representations. The results are evaluated over 500 episodes in 500 unseen maps.}
    \label{fig:absolute_relative}
    \vspace{-8pt}
\end{figure}

As shown in Fig.~\ref{fig:absolute_relative}, Roken with absolute trajectory representation fails to learn meaningful multi-robot coordination distributions.
The full success ratio is only $5.3\%$, compared with $75.8\%$ for the relative representation.
Although we have normalized the absolute positions to a fixed range, the model still struggles to learn.
The reason is that the absolute trajectory representation buries the underlying coordination patterns across different samples by applying different initial position offsets.
For example, two identical episodes with a slight shift of the map and trajectories will be treated as two distinct samples with the absolute representation, while they share exactly the same relative movements.
Therefore, we use the relative trajectory representation in all our experiments to facilitate the learning of multi-robot coordination patterns.

Note that in all our experiments, we use the normalized absolute position for robot token position encoding, rather than the relative position representation w.r.t. the leader robot.
With this setting, we found that each token can successfully cross-attend to the relevant map information and other robots' states to generate coordinated trajectories.
It further simplifies the application of the Roken model without the need of additional relative position transformation.

\subsection{Implementations of Masked Training and Clustered Rollout}

The clustered rollout is supported by the masked training of Roken, where the model predicts the trajectories of a subset of robots conditioned on the previously generated trajectories of the other robots.
With this, we can scale the pre-trained Roken to even larger team sizes by dividing the team into overlapped clusters and performing rollout for each cluster sequentially.
The trajectories of the overlapped robots are kept fixed as conditions.

\begin{figure}[h]
    \centering
    \includegraphics[width=.7\linewidth]{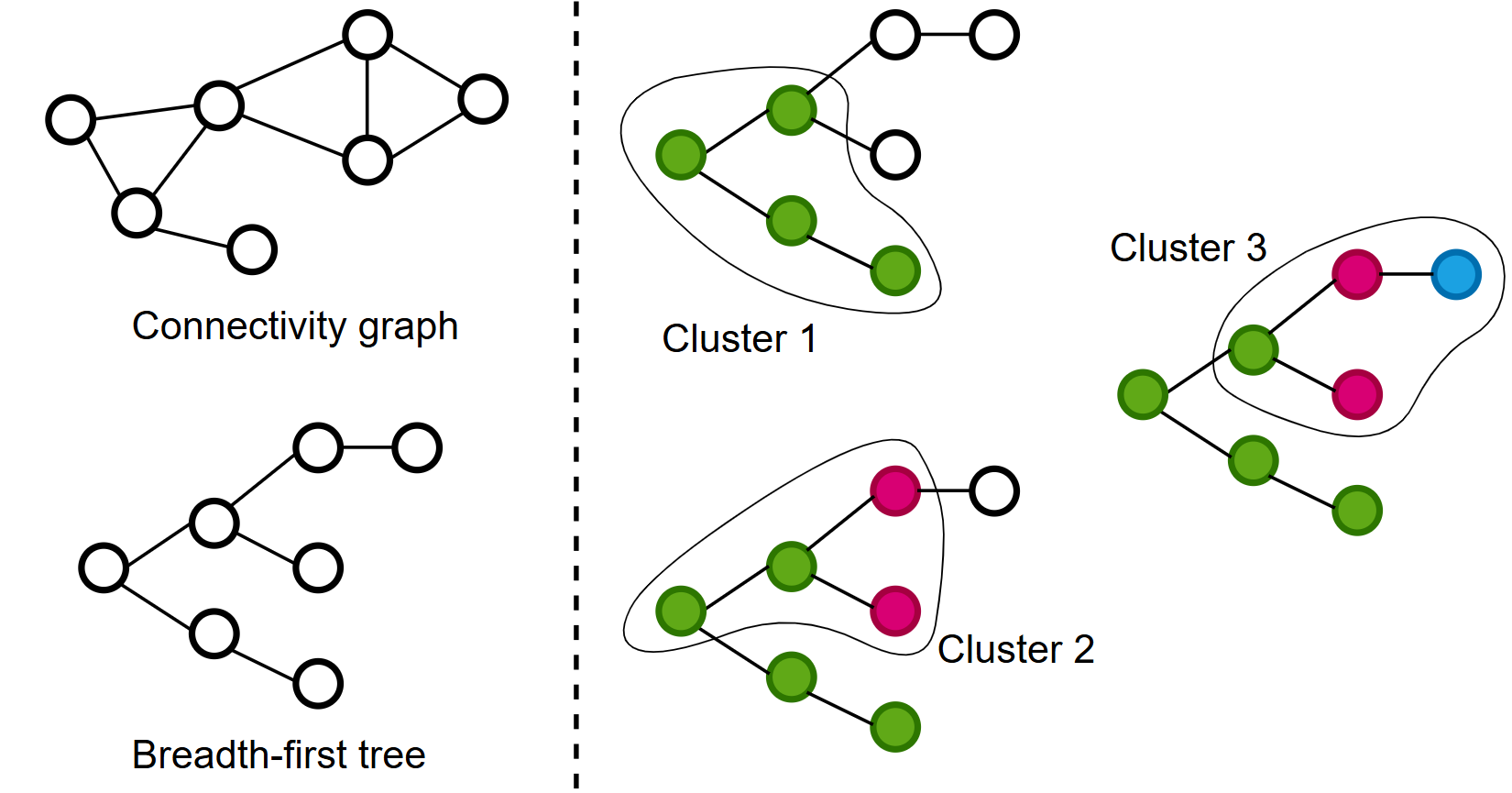}
    \caption{Example of clustered rollout. After Breadth-first tree construction, we cluster the robots into overlapping groups, which can then be generated sequentially. With masked training of Roken and clustered rollout, we can scale the pre-trained Roken only on the four robot dataset to 7 robots without additional training.}
    \label{fig:clustered_rollout}
\end{figure}

\textbf{Overlapped clustering:}
As shown in Fig.~\ref{fig:clustered_rollout}, the overlapped clustering of the team includes three steps: 
(1) communication graph construction, as introduced in Sec.~\ref{sec:communication_graph}; 
(2) BFS tree construction with the leader robot as the root; and (3) cluster construction from the BFS tree.
In cluster construction, the first cluster is rooted at the leader robot, and includes its children in the BFS tree until reaching the cluster size limit (which we set as 4).
Starting from the second cluster, we iteratively select a robot that has been generated in the previous clusters as the root, and construct a new cluster with its children until reaching the cluster size limit. If the unvisited children are not enough to fill the cluster, we will further expand the cluster with the siblings of newly added children, and further the siblings or ancestors of the root robot until reaching the cluster size limit. 
There are three possible conditional generation cases: (1) one parent condition robot token with three unvisited children tokens; (2) two conditional robot tokens with two unvisited children tokens; and (3) three conditional robot tokens with one unvisited child token.
We bias the case (3) by adding visited siblings or ancestors to fill the cluster, which can provide more informative conditions for the generation of the unvisited child token, at the cost of more clusters to cover the full team.

Clustered rollout can scale the pre-trained Roken to larger team sizes without additional training, at the cost of more generation steps and thus more compounding errors.
The scalability performance of the clustered rollout is shown in Fig.~\ref{fig:scalability}, where the pre-trained Roken$_{\text{four}}^{\text{masked}}$ can be applied to even ten robots with a full success ratio of $68.4\%$, in contrast to the $10.5\%$ success ratio of Roken$_{\text{four}}$ without masked training.

\subsection{Ablations on Replanning Frequency}
\begin{figure}[h]
    \centering
    \includegraphics[width=.5\linewidth]{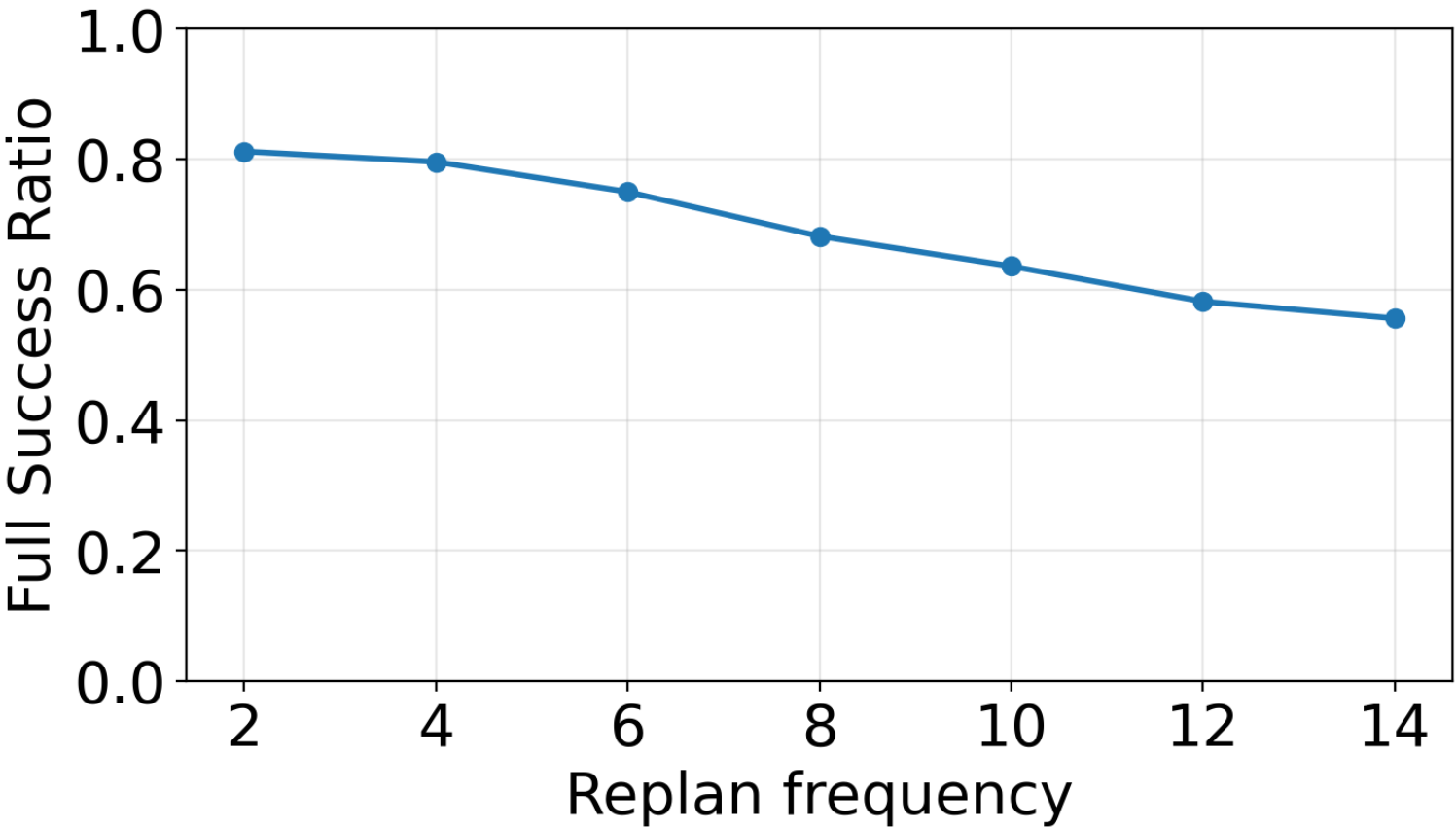}
    \caption{Ablation on replanning frequency. The model Roken$_{\text{four}}$ is rollout in 500 episodes with 500 unseen maps. The robot number is fixed to four.}
    \label{fig:replanning_frequency}
\end{figure}

In both our simulation rollout and real-world deployment, we adopt a receding horizon control scheme where Roken generates a trajectory chunk with 16 steps at every control cycle, but only executes the first two steps before re-planning with the updated multi-robot state and occupancy map.
Here we provide results of Roken with different replanning frequencies by changing the number of executed steps before the next re-planning, as shown in Fig.~\ref{fig:replanning_frequency}.
The full success ratio decreases as the replanning frequency decreases, which indicates that more frequent replanning can help to correct errors in the predicted trajectory with more updated states. 
Therefore, we set the replanning frequency to 2 steps in all our evaluations to get the best performance.

\subsection{Sim-to-Real Gap and Real-World Deployment}

To deploy the trained Roken to real-world multi-robot systems, we need to consider the sim-to-real gap, especially how to apply Roken (trained with fixed parameters) to different communication and safety requirements.

\begin{itemize}
    \item \textbf{Obstacle avoidance with different robot sizes:} This constraint can be addressed by adjusting the expansion radius of the obstacles in the occupancy map.
    \item \textbf{Communication maintenance with different radius:} This constraint can be addressed by scaling the local occupancy maps with different resolutions. The pre-defined communication range is 15 pixels in the input occupancy map where 1 pixel represents 1 meter. Therefore, shrinking the original $\mathcal{H}_{\mathrm{map}}$ to a smaller size will result in a larger communication range, while expanding the original $\mathcal{H}_{\mathrm{map}}$ will result in a smaller communication range (as one pixel now represents less than 1 meter).
    \item \textbf{Inter-robot collision avoidance:} The current Roken cannot directly adapt to different inter-robot collision distances. In our real-world deployment, this limitation is not a problem because the \emph{lazy movement} behavior of the trained Roken policy encourages robots to stay still when possible to reduce the efforts for connectivity maintenance. Therefore, the robots tend to form a chain-like formation in long-term navigation tasks. To strictly impose a different inter-robot collision distance, a safety shield module is needed to perform online trajectory correction when the predicted trajectory violates the inter-robot collision constraint.
\end{itemize}

For real-world deployment, we use four WheelTec robots\footnote{https://wheeltec.net/} with omnidirectional mecanum wheels.
The localization of each robot is obtained by the AprilTag-based system\footnote{https://github.com/AprilRobotics/apriltag}, and the occupancy map provided as a prior.
The robots track the generated trajectories with a simple PI controller with a control frequency of 100Hz.
The deployment results of one to four robots are shown in Fig.~\ref{fig:real_deployment}.
The results are also included in the supplementary video.

\begin{figure}
    \centering
    \includegraphics[width=\linewidth]{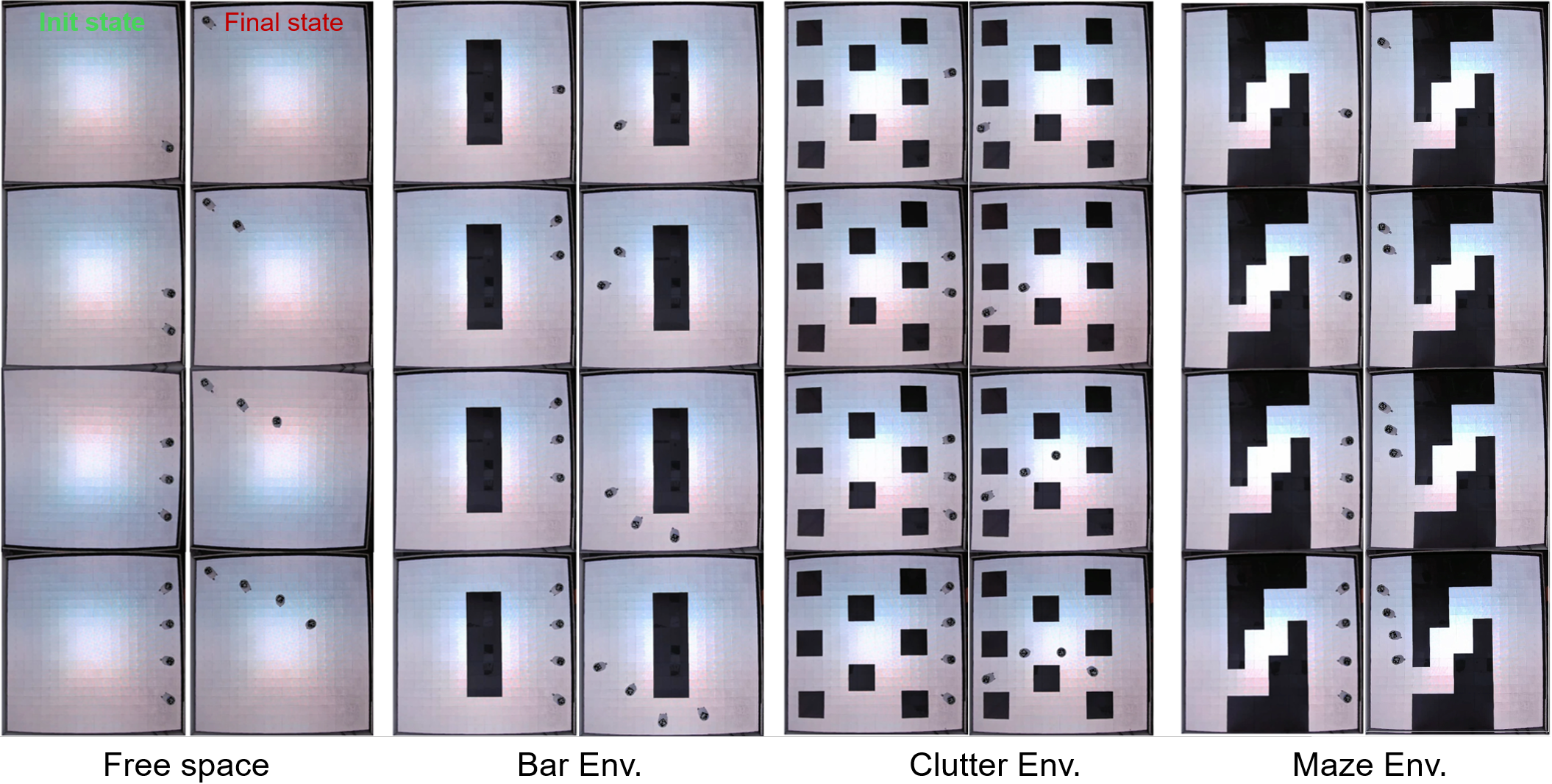}
    \caption{Real-world deployment of Roken. For each environment, the left column shows the initial state and the right column shows the final state after navigation. One Roken model can successfully control one to four robots to navigate to the goal while maintaining connectivity and avoiding collisions.}
    \label{fig:real_deployment}
\end{figure}

\subsection{More Rollout Results}

\begin{figure}[h]
    \centering
    \includegraphics[width=\linewidth]{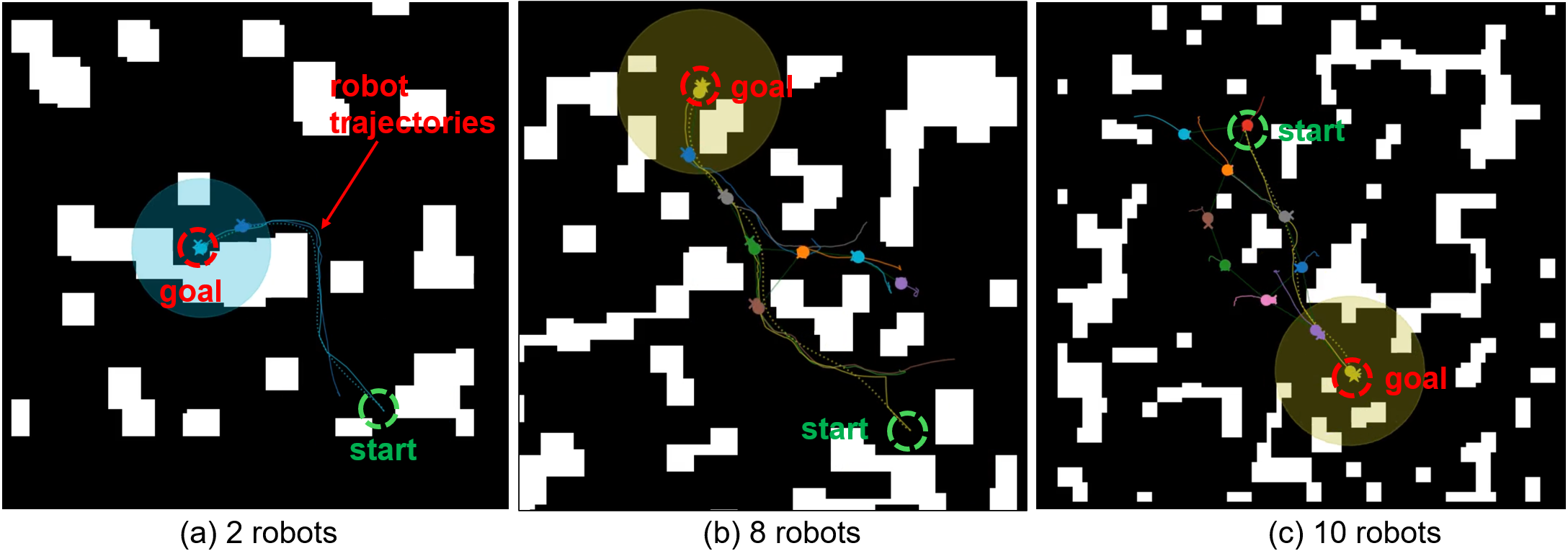}
    \caption{Rollout episodes of Roken$_{\text{mixed}}$ for 2, 8, and 10 robots.}
    \label{fig:more_rollout_results}
\end{figure}

\begin{figure}[h]
    \centering
    \includegraphics[width=\linewidth]{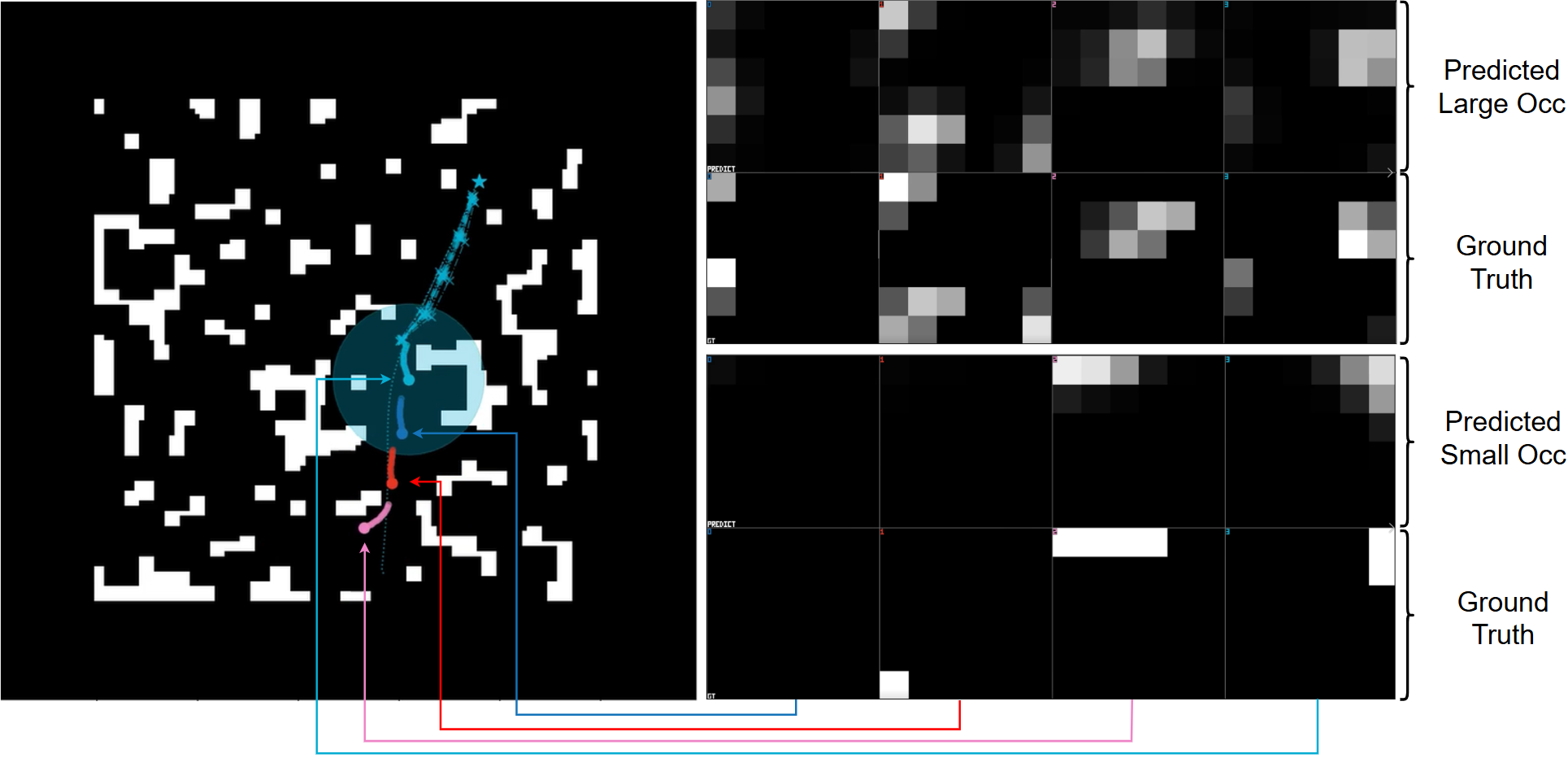}
    \includegraphics[width=\linewidth]{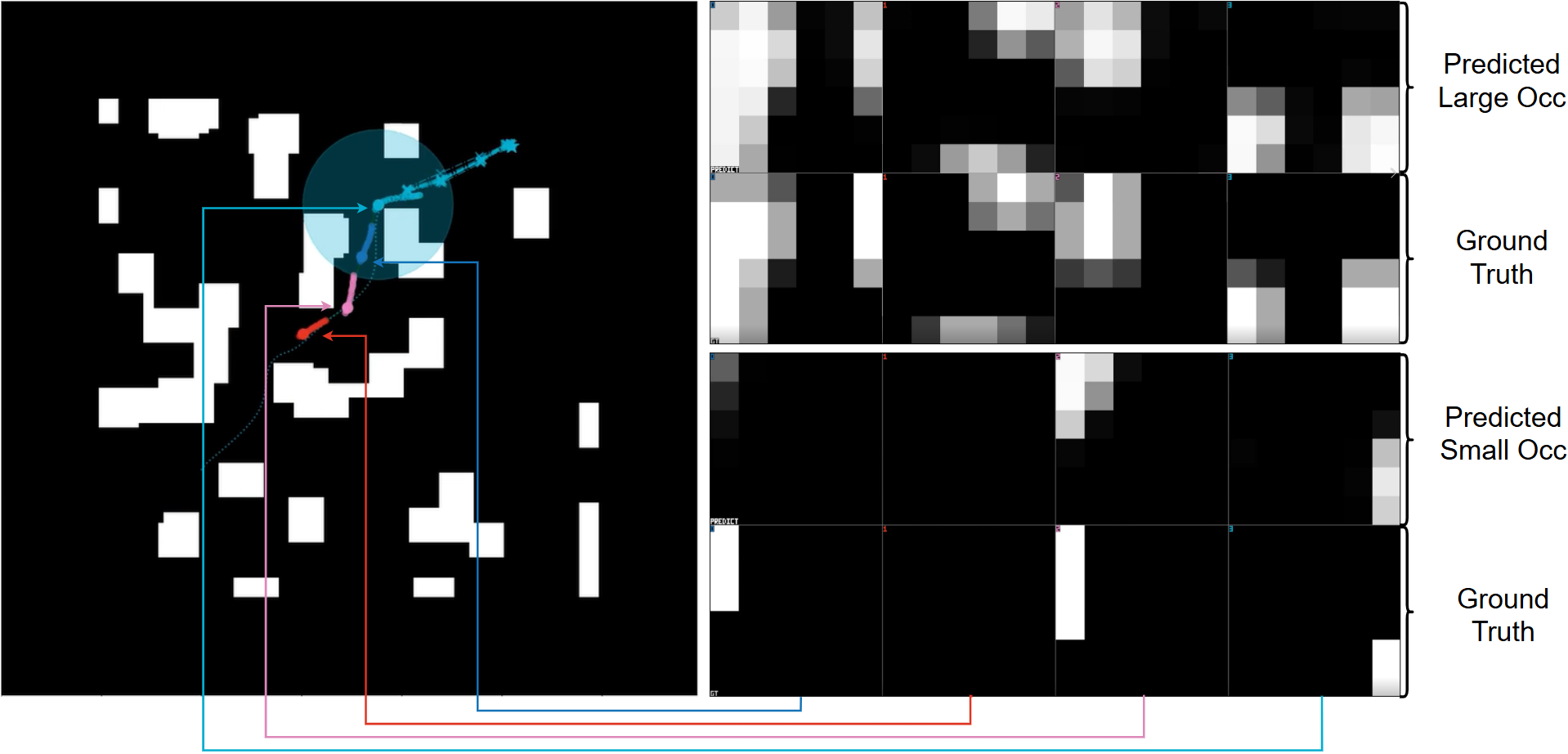}
    \caption{Examples of both large and small local occupancy reconstruction results. Each row corresponds to one robot. The predicted occupancy patches demonstrate that in Roken, robots can successfully attend to multiple discrete map tokens, aggregate them, and reconstruct their surrounding local occupancy pattern.}
    \label{fig:occ_reconstruction}
\end{figure}

\begin{figure}[!t]
    \centering
    \includegraphics[width=\linewidth]{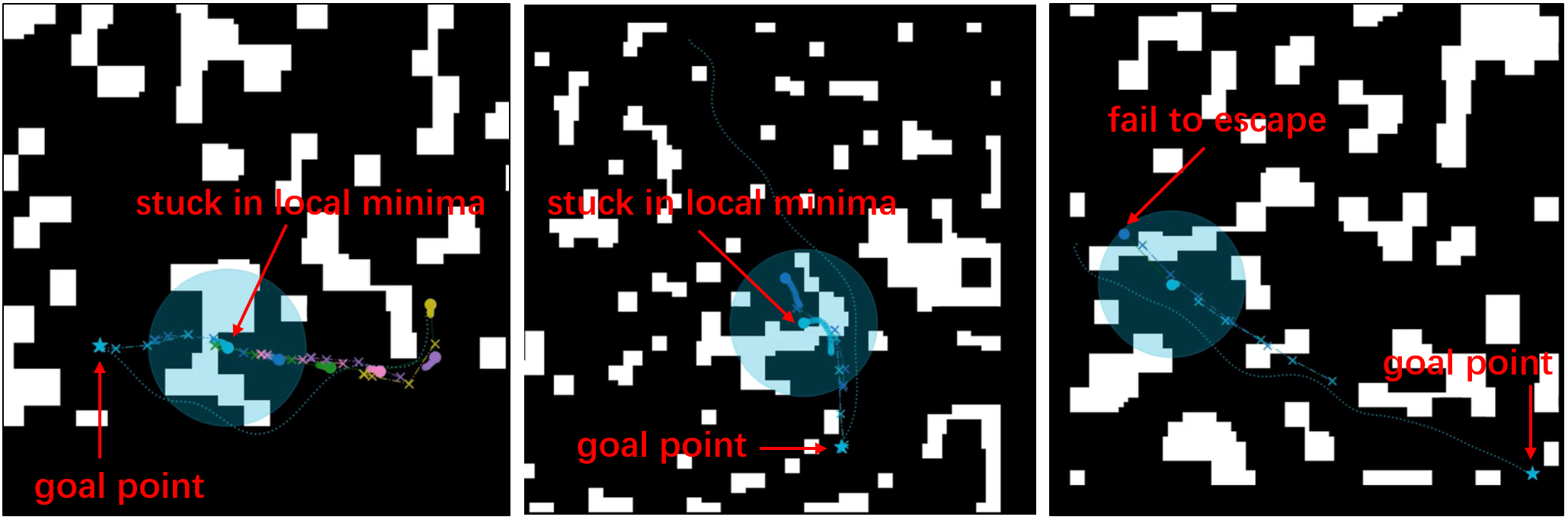}
    \caption{Example failure cases of Roken. Roken fails to generate valid trajectories that escape from the local minima areas, because the training dataset does not include expert demonstrations in such scenarios. This can be further improved by dataset enhancement and RL post-training.}
    \label{fig:failure_cases}
\end{figure}

\end{document}